\pdfoutput=1

\documentclass[11pt]{article}

\usepackage[preprint]{acl}

\usepackage{times}
\usepackage{latexsym}

\usepackage[T1]{fontenc}

\usepackage[utf8]{inputenc}

\usepackage{microtype}

\usepackage{inconsolata}

\usepackage{graphicx}

\newcommand{\troltitle}{\includegraphics[width=0.07\textwidth, trim=0in 2in 0in 0in]{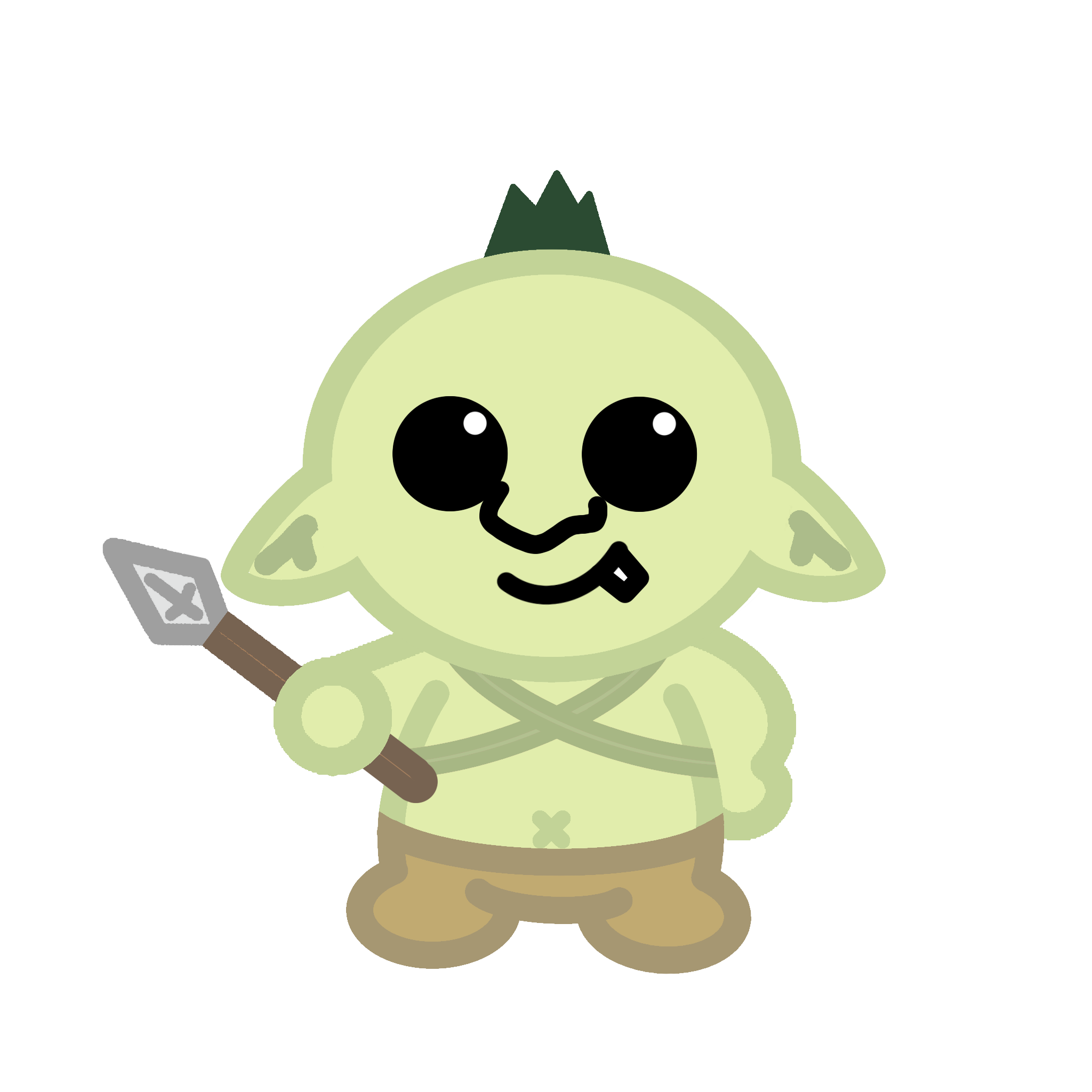}\hspace{-0.3ex}} 
\newcommand{\trol}{\includegraphics[width=0.029\textwidth, trim=0in 2in 0in 0in]{figures/trol_emoji.png}\hspace{-0.2ex}} 
\usepackage{float}
\usepackage{graphicx}
\usepackage{booktabs} 
\usepackage{multirow}
\usepackage{amsmath}
\usepackage{xcolor} 
\usepackage{caption}
\usepackage{pifont} 
\usepackage{color, colortbl} 
\usepackage{arydshln} 
\usepackage{tabularx} 
\usepackage{footnote} 
\usepackage{amsfonts} 

\title{\troltitle TroL: Traversal of Layers for Large Language and Vision Models}

\author{Byung-Kwan Lee \\
  KAIST \\
  \texttt{ \small \centering leebk@kaist.ac.kr} \\
  \And 
  Sangyun Chung \\
  KAIST \\
  \texttt{\small \centering  jelarum@kaist.ac.kr} \\
  \And
  Chae Won Kim \\
  KAIST \\
  \texttt{\small \centering  chaewonkim@kaist.ac.kr} \\
  \AND
  Beomchan Park \\
  KAIST \\
  \texttt{\small \centering  bpark0810@kaist.ac.kr} \\
  \And
  Yong Man Ro\\
  KAIST\\
  \texttt{\small \centering  ymro@kaist.ac.kr}
  }

\begin{document}
\maketitle

\begin{abstract}
Large language and vision models (LLVMs) have been driven by the generalization power of large language models (LLMs) and the advent of visual instruction tuning. Along with scaling them up directly, these models enable LLVMs to showcase powerful vision language (VL) performances by covering diverse tasks via natural language instructions. However, existing open-source LLVMs that perform comparably to closed-source LLVMs such as GPT-4V are often considered too large (\textit{e.g.}, 26B, 34B, and 110B parameters), having a larger number of layers. These large models demand costly, high-end resources for both training and inference. To address this issue, we present a new efficient LLVM family with 1.8B, 3.8B, and 7B LLM model sizes, \textbf{Tr}aversal \textbf{o}f \textbf{L}ayers (\trol \textbf{TroL}), which enables the reuse of layers in a token-wise manner. This layer traversing technique simulates the effect of looking back and retracing the answering stream while increasing the number of forward propagation layers without physically adding more layers. We demonstrate that \trol \textbf{TroL} employs a simple layer traversing approach yet efficiently outperforms the open-source LLVMs with larger model sizes and rivals the performances of the closed-source LLVMs with substantial sizes. Code is available in \href{https://github.com/ByungKwanLee/TroL}{https://github.com/ByungKwanLee/TroL}.
\end{abstract}
\section{Introduction}
\label{sec:intro}

\begin{figure}[t!]
\vspace{0mm}
    \centering
    \includegraphics[width=0.3\textwidth]{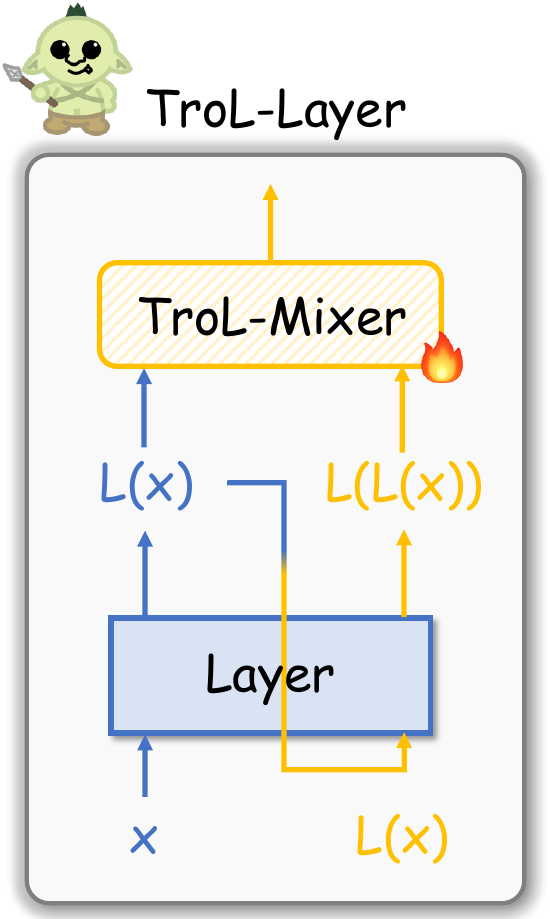}
    \vspace{-2mm}
    \caption{Overview of \trol TroL's layer traversing}
    \label{fig:figure1}
    \vspace{-5mm}
\end{figure}

\begin{figure*}[t!]
\vspace{-6mm}
    \centering
    \includegraphics[width=\textwidth]{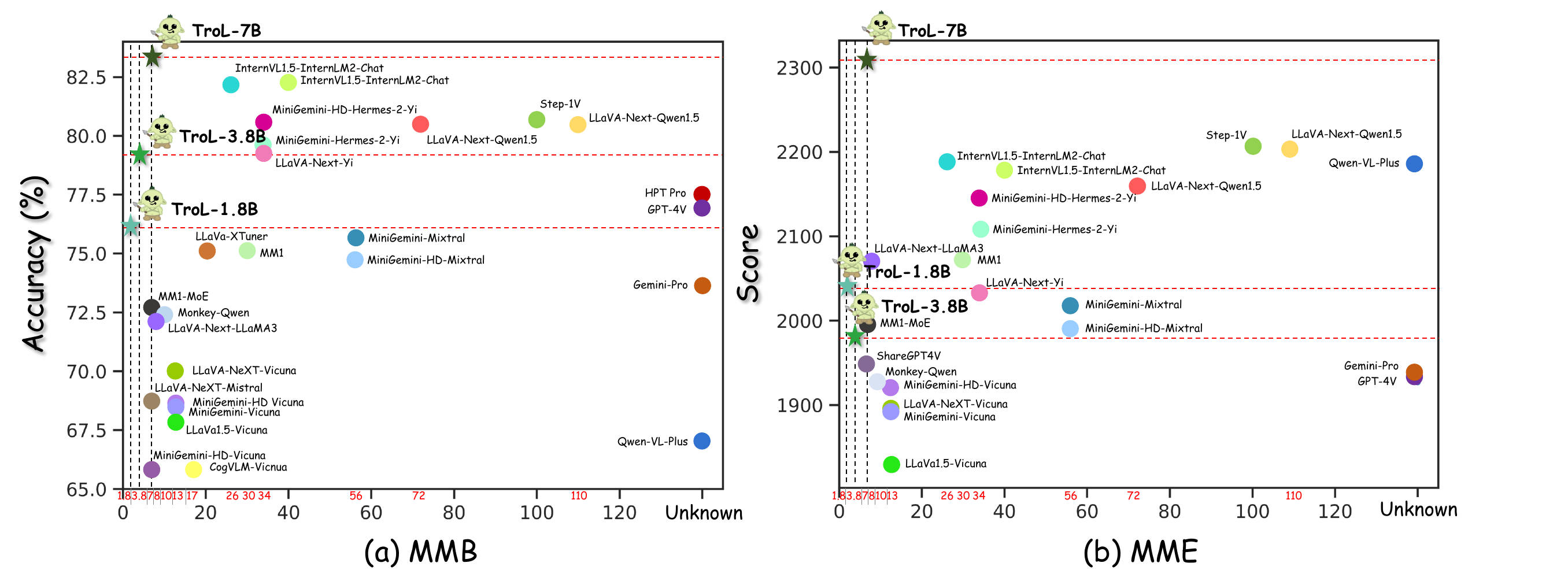}
    \vspace{-9mm}
    \caption{Performances for efficient LLVM family, \trol TroL, across three model sizes: 1.8B, 3.8B, and 7B}
    \label{fig:figure2}
    \vspace{-5mm}
\end{figure*}

The great success of closed-source large language and vision models (LLVMs) such as GPT-4V~\citep{achiam2023gpt}, Gemini-Pro~\citep{team2023gemini}, and Qwen-VL-Plus~\citep{bai2023qwen} has prompted the world to build open-source LLVMs using open-source large language models (LLMs) and visual instruction tuning~\citep{liu2023visual, liu2023improved, liu2024llavanext}. These movements are expected to contribute significantly to both research and industry through numerous downstream tasks, such as on-device chatbot systems.

In order to bootstrap their vision language performance, several studies have curated high-quality visual instruction tuning datasets~\citep{chen2023sharegpt4v, chen2024allava} by leveraging the power of closed-source LLVMs. Further, they have physically increased the model sizes~\citep{mckinzie2024mm1, li2024mini, liu2024llavanext} to enhance their capability to understand complex question-answer pairs.

Following these efforts, there has been a recent rise in techniques for employing additional modules: mixed vision encoders~\citep{kar2024brave, lu2024deepseek, goncharova2024omnifusion, ranzinger2023radio, zhao2024cobra}, multiple computer vision models~\citep{chen2024spatialvlm, wang2024all, jiao2024enhancing, lee2024collavo, lee2024moai}, and mixtures of experts (MoE) for efficiently scaling LLVMs to fulfill specific purposes~\citep{lin2024moe, lee2024moai, li2024cumo, li2024uni, guo2024dynamic, mckinzie2024mm1, li2024mini, gao2024sphinx, sun2024parrot}. In the end, LLVMs combined with these techniques have shown improved vision language performances, achieving results comparable to directly scaled-up LLVMs with 26B, 34B, and 110B model sizes. Additionally, some approaches have demonstrated performances that surpass those of the closed-source LLVMs.

\begin{figure*}[t!]
\vspace{-5mm}
    \centering
    \includegraphics[width=\textwidth]{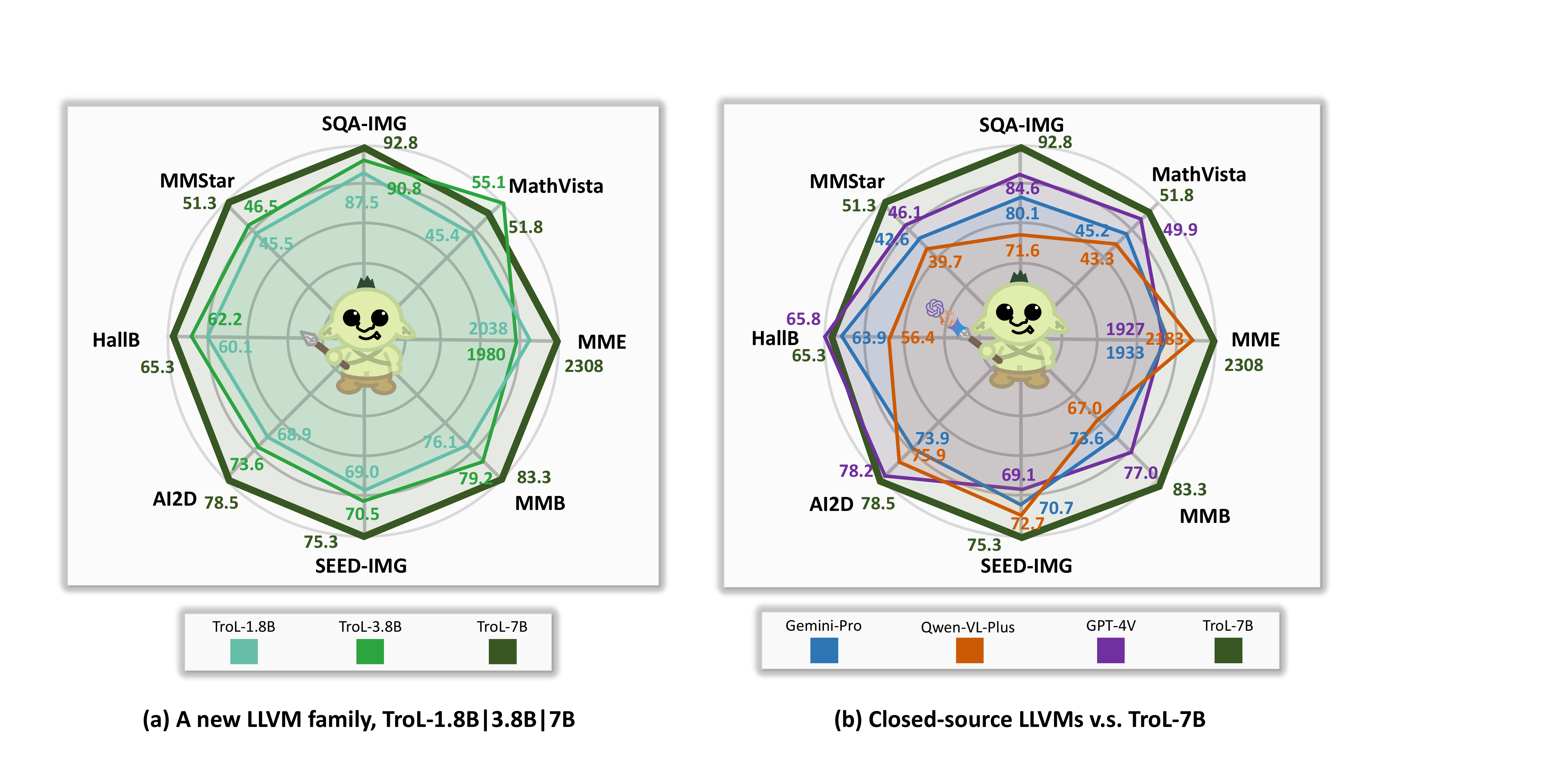}
    \vspace{-5mm}
    \caption{Overview of vision language performances compared with the \trol TroL and closed-source LLVMs}
    \label{fig:figure3}
    \vspace{-5mm}
\end{figure*}

However, directly scaling the model size up or using additional modules may not be considered a fundamental solution to enlarging learning capabilities regarding complex question-answer pairs. This is because they physically add a considerable number of training parameters or borrow richer knowledge from external modules. In other words, it remains unexplored how LLVMs with smaller model sizes can effectively enhance learning capabilities despite their inherent physical limitations.

To address this, simply increasing the image resolution size~\citep{li2023otterhd, bai2023qwen, wang2023cogvlm, ye2023mplug2, hu2024mplug} and dynamically dividing images into sub-parts for hierarchical focus~\citep{liu2024llavanext, mckinzie2024mm1, xu2024llava} may be good candidates to achieve our purpose without employing any additional modules mentioned. Nonetheless, these strategies are mostly intended to embed rich image information to further improve overall and fine-grained image understanding. Hence, we need to focus on how multimodal LLMs can be enhanced by themselves. Once more efficient LLMs that perform comparably to closed-source models are released publicly, it will mitigate the necessity for high-end GPUs and accelerate significant advancements in various downstream applications, including on-device processing.

Therefore, we present a new efficient LLVM family with 1.8B, 3.8B, and 7B model sizes, \textbf{Tr}aversal \textbf{o}f \textbf{L}ayers (\trol\textbf{TroL}), which enables the reuse of layers in a token-wise manner. To overcome the inherent limitations of smaller-sized LLVMs, we opt to increase the number of forward propagations rather than physically adding more layers, as is normally done in scaled-up LLVMs. This technique, which we call layer traversing, allows LLVMs to retrace and re-examine the answering stream, akin to human retrospection and careful thought before responding with an answer. Figure~\ref{fig:figure1} represents how the layer traversing technique is practically implemented in TroL-Layer, where TroL-Mixer serves as the token-wise mixing operation under lightweight additional parameters: 49K, 98K, and 131K in total layers. This is a significantly tiny number compared with the 1.8B, 3.8B, and 7B model sizes.

To successfully apply layer traversing to LLVMs, we employ a two-step training process, establishing \trol\textbf{TroL}. The first step involves training a vision projector and all TroL-Mixers for each TroL-Layer. This is a crucial step because it not only aligns vision and language information but also tunes the TroL-Mixers with the answering stream in backbone multimodal LLMs, thereby facilitating the use of layer traversing. The second training step includes further training of these components along with the backbone multimodal LLMs. To achieve efficient training, we use Q-LoRA~\citep{dettmers2023qlora} training for the backbone multimodal LLMs under 4/8-bit quantization.

In conducting the two-step training, we demonstrate that \trol\textbf{TroL} is an efficient model, yet it outperforms open-source LLVMs with larger model sizes (e.g., 26B, 34B, 72B, and 110B) and closed-source LLVMs with a substantially vast amount of parameters, as illustrated in Figures~\ref{fig:figure2} and \ref{fig:figure3}.

Our contribution can be summarized into two main aspects:
\begin{itemize}
\item We introduce a new efficient LLVM family—1.8B, 3.8B, and 7B, \textbf{Tr}aversal \textbf{o}f \textbf{L}ayers (\trol\textbf{TroL}), which enables the reuse of layers, simulating the effect of retracing the answering stream.
\item \trol\textbf{TroL} proves its superior effectiveness on various evaluation benchmarks compared with substantially sized open- and closed-source LLVMs without directly scaling up the model size and without any additional modules.
\end{itemize}

\section{Related Works}
\label{sec:related}

\begin{figure*}[t!]
\vspace{-5mm}
    \centering
    \includegraphics[width=0.9\textwidth]{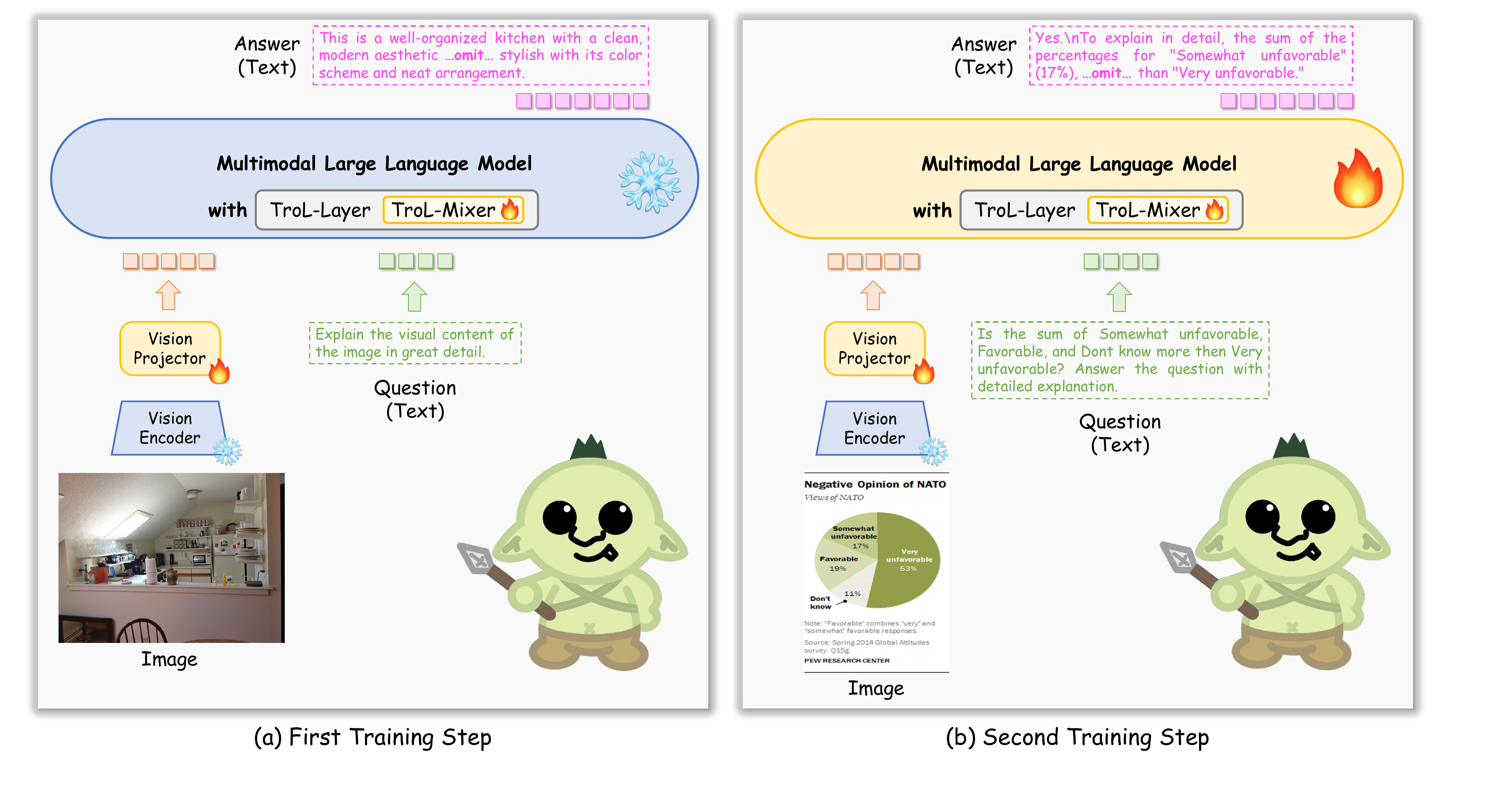}
    \vspace{-3mm}
    \caption{Overview of two-step training to build an efficient LLVM Family, \trol TroL}
    \label{fig:figure4}
    \vspace{-5mm}
\end{figure*}

\paragraph{Large Language and Vision Models.}

The curation of visual instruction tuning datasets ~\citep{liu2023visual, liu2023improved, liu2024llavanext, dai2023instructblip} has significantly propelled the rapid development of LLVMs \citep{chen2023shikra, bai2023qwen, zhu2023minigpt, li2023otter, ye2023mplug, ye2023mplug2, chen2023sharegpt4v, 2023xtuner, zhang2023internlm, chen2023internvl, chen2024far}. Building upon this foundation, recent research has further advanced LLVMs' capabilities through two key strategies: scaling up model sizes and designing specialized instruction tuning datasets. Firstly, increasing model sizes has emerged as a prominent approach to enhancing LLVM performance and capacity. For example, LLVMs \citep{mckinzie2024mm1, li2024mini, liu2024llavanext, wang2023cogvlm, laurenccon2023obelisc, sun2023generative, gao2024sphinx, sun2024parrot} deploy larger architectures and more parameters to enlarge the representational power of these models. Moreover, the development of meticulous instruction tuning datasets \citep{chen2023sharegpt4v, li2024mini, hu2024mplug, gao2023g, wang2024measuring, yue2023mammoth, yue2024mammoth2} has played a pivotal role in improving LLVMs for specific tasks or domains. 

Additionally, recent research has explored more direct approaches to enhance LLVM’s image perception capabilities, either by modifying visual input and utilizing additional modules. For example, Qwen-VL \citep{bai2023qwen}, CogVLM \citep{wang2023cogvlm}, and mPLUG family \citep{ye2023mplug2, hu2024mplug} increase image resolution to enrich visual information in LLVMs. Moreover, few approaches \citep{liu2024llavanext, mckinzie2024mm1, li2024mini} improve visual tokens through image partitioning. Additionally, the integration of additional vision encoders \citep{kar2024brave, lu2024deepseek, goncharova2024omnifusion, ranzinger2023radio, zhao2024cobra} and external computer vision modules \citep{chen2024spatialvlm, wang2024all, jiao2024enhancing, lee2024collavo, lee2024moai} augment LLVMs' image perception capability, thereby enhancing overall performance on multimodal tasks.

While these lines of works expand the overall learning capabilities of LLVMs, they do not necessarily enhance the fundamental capacity of LLVMs. Consequently, there remains a need for further research into the intrinsic mechanisms of LLVMs without scaling models or leveraging additional modules. Thus, we introduce a new LLVM family, \trol TroL, specifically tailored to enhance the learning capabilities of LLVMs, where layer traversing is presented to efficiently reuse layers. This approach simulates the effect of retracing the answering stream, offering a focused solution to propel the advancement of LLVMs.

\paragraph{Numerous Efficient Approaches.}
Despite the remarkable achievements of LLVMs in a short period of time, closed-source LLVMs require a significant number of parameters and resources. For that reason, many LLVMs have been exploring methods to create more efficient models. These methods aim to reduce the number of parameters or improve computational efficiency without significantly compromising performance. Numerous methods to reducing the number of parameters involve sharing specific weights within the model \citep{thawakar2024mobillama, lan2020albert, takase2023lessons, reid2021subformer}, eliminating weights that contribute less to the performance \citep{sun2024simple, ma2023llmpruner, cao2023pumer, frantar2023sparsegpt, men2024shortgpt}, and employing quantization techniques \citep{shao2024omniquant, li2023loftq, park2024lutgemm}.

These efforts primarily focus on reducing inference time or accelerating training speed while maintaining performance, rather than fundamentally improving it. In contrast, \trol TroL aims to efficiently enhance the learning capabilities of LLVMs in understanding complex question-answer pairs.

\definecolor{Gray}{gray}{0.93}
\definecolor{Green}{rgb}{0.8509, 0.952, 0.717}
\newcommand{\cmark}{\ding{51}}%
\newcommand{\xmark}{\ding{55}}%

\begin{table*}[t!]
\vspace{-6mm}
\centering
\resizebox{\linewidth}{!}{
\renewcommand{\tabcolsep}{0.8mm}
\begin{tabular}{lccccccccccccc}
\toprule
LLVMs     & Q-Bench & SQA$^{\text{I}}$ & AI2D & ChartQA & SEED$^{\text{I}}$ & POPE & HallB & MME & MathVista & MMB & MMB$^{\text{CN}}$ & MM-Vet & LLaVA$^{\text{W}}$  \\
\midrule
BLIP2-13B~\cite{blip2}                            & -    & 61.0 & -    & -    & 46.4 & 85.3 & -    & 1584 & -    & -    & -    & 22.4 & -    \\
InstructBLIP-7B~\cite{dai2023instructblip}        & 56.7 & 60.5 & -    & -    & 53.4 & -    & 53.6 & -    & 25.3 & 36.0 & 23.9 & 26.2 & -    \\
InstructBLIP-13B~\cite{dai2023instructblip}       & -    & 63.1 & -    & -    & -    & 78.9 & -    & -    & -    & 33.9 & -    & 25.6 & -    \\
IDEFICS-9B~\cite{laurenccon2023obelisc}           & 51.5 & -    & -    & -    & -    & 74.6 & -    & 1353 & 19.8 & 48.2 & 25.2 & 23.7 & -    \\
Qwen-VL-7B~\cite{bai2023qwen}                     & 59.4 & 67.1 & -    & -    & -    & -    & -    & -    & -    & 38.2 & 7.4  & -    & -    \\
Qwen-VL-Chat-7B~\cite{bai2023qwen}                & 33.8 & 68.2 & -    & -    & 58.2 & -    & 56.4 & 1849 & -    & 60.6 & 56.7 & 47.3 & -    \\
MiniGPT-4-7B~\cite{zhu2023minigpt}                & 51.8 & -    & -    & -    & -    & -    & -    & -    & 23.1 & 23.0 & 11.9 & 22.1 & -    \\
Otter-7B~\cite{li2023otter}                       & 47.2 & -    & -    & -    & -    & 72.5 & -    & 1599 & 19.7 & 48.3 & -    & 24.7 & -    \\
UIO-2-XXL-6.8B~\cite{lu2023unifiedio}              & -    & 86.2 & -    & -    & 61.8 & 87.7 & -    & -    & -    & 71.5 & -    & -    & -       \\
\rowcolor{Gray}
LLaVA-7B~\cite{liu2023visual}                     & -    & 38.5 & -    & -    & -    & 80.2 & 44.1 & 1055 & -    & 34.1 & 14.1 & 26.7 & -    \\
\rowcolor{Gray}
LLaVA1.5-7B~\cite{liu2023improved}                & 60.1 & 66.8 & -    & -    & 58.6 & 85.9 & -    & 1805 & -    & 64.3 & 58.3 & 30.5 & 63.4 \\
\rowcolor{Gray}
LLaVA1.5-13B~\cite{liu2023improved}               & 61.4 & 71.6 & 54.8 & 18.2 & 61.6 & 85.9 & 46.7 & 1826 & 27.6 & 67.7 & 63.6 & 35.4 & -    \\
mPLUG-Owl-7B~\cite{ye2023mplug}                   & 58.9 & -    & -    & -    & -    & -    & -    & -    & 22.2 & 46.6 & -    & -    & -    \\
mPLUG-Owl2-7B~\cite{ye2023mplug2}                 & 62.9 & 68.7 & -    & -    & -    &      & -    & -    & -    & 64.5 & 60.3 & 36.2 & -    \\
ShareGPT4V-7B~\cite{chen2023sharegpt4v}           & 63.4 & 68.4 & -    & -    & 69.7 & -    & 49.8 & 1944 & 25.8 & 68.8 & 62.2 & 37.6 & -    \\
InternLM-XC-7B~\cite{zhang2023internlm}           & 64.4 & -    & -    & -    & 66.1 & -    & 57.0 & 1919 & 29.5 & 74.4 & 72.4 & 35.2 & -    \\
Monkey-10B~\cite{li2023monkey}                    & -    & 69.4 & -    & -    & 68.9 & -    & 58.4 & 1924 & 34.8 & 72.4 & 67.5 & 33.0 & -    \\
VILA-7B~\cite{lin2023vila}                        & -    & 68.2 & -    & -    & 61.1 & 85.5 & -    & -    & -    & 68.9 & 61.7 & 34.9 & -    \\
VILA-13B~\cite{lin2023vila}                       & -    & 73.7 & -    & -    & 62.8 & 84.2 & -    & -    & -    & 70.3 & 64.3 & 38.8 & -    \\
SPHINX-7B~\cite{lin2023sphinx}                    & -    & 70.6 & -    & -    & 71.6 & 86.9 & -    & 1797 & 27.8 & 65.9 & 57.9 & 40.2 & -    \\
SPHINX-MoE-7B$\times$8~\cite{gao2024sphinx}       & 66.2 & 70.6 & -    & -    & 73.0 & \textbf{89.6} & -    & 1852 & 42.7 & 71.3 & -    & 40.9 & -    \\
SPHINX-Plus-13B~\cite{gao2024sphinx}              & 66.2 & 70.6 & -    & -    & 74.8 & 89.1 & 52.1 & 1741 & 36.8 & 71.0 & -    & 47.9 & -    \\
\rowcolor{Gray}
LLaVA-NeXT-7B~\cite{liu2024llavanext}             & -    & 70.1 & -    & -    & 70.2 & 86.5 & -    & 1851 & 34.6 & 69.6 & 63.3 & 43.9 & 72.3 \\
\rowcolor{Gray}
LLaVA-NeXT-8B~\cite{liu2024llavanext}             & -    & -    & 71.6 & 69.5 & -    & -    & -    & 1972 & 37.5 & 72.1 & -    & -    & 80.1 \\
\rowcolor{Gray}
LLaVA-NeXT-13B~\cite{liu2024llavanext}            & -    & 73.6 & 70.0 & 62.2 & 72.2 & 86.7 & -    & 1892 & 35.1 & 70.0 & 68.5 & 47.3 & 72.3 \\
MM1-7B~\cite{mckinzie2024mm1}                     & -    & 72.6 & -    & -    & 69.9 & 86.6 & -    & 1858 & 35.9 & 72.3 & -    & 42.1 & -    \\
MM1-MoE-7B$\times$32~\cite{mckinzie2024mm1}       & -    & 74.4 & -    & -    & 70.9 & 87.8 & -    & 1992 & 40.9 & 72.7 & -    & 45.2 & -    \\
MiniGemini-HD-7B~\cite{li2024mini}                & -    & -    & -    & -    & -    & -    & -    & 1865 & 32.2 & 65.8 & -    & 41.3 & -    \\
MiniGemini-HD-13B~\cite{li2024mini}               & -    & -    & -    & -    & -    & -    & -    & 1917 & 37.0 & 68.6 & -    & 50.5 & -    \\
\midrule
\rowcolor{Green}
TroL-7B     & \textbf{73.6} 
            & \textbf{92.8}
            & \textbf{78.5}    
            & \textbf{71.2}
            & \textbf{75.3} 
            & 87.8
            & \textbf{65.3}    
            & \textbf{2308}
            & \textbf{51.8}
            & \textbf{83.5} 
            & \textbf{81.2}  
            & \textbf{54.7} 
            & \textbf{92.8}    \\
\bottomrule 
\end{tabular}
}
\vspace{-2mm}
\caption{Comparison with the current existing standard model size open-source LLVMs, evaluating vision language performances of \trol TroL on numerous evaluation benchmarks: Q-Bench~\cite{wu2023q}, SQA$^{\text{I}}$~\cite{lu2022learn}, AI2D~\cite{kembhavi2016diagram}, ChartQA~\cite{masry2022chartqa}, SEED$^{\text{I}}$~\cite{li2023seed}, POPE~\cite{li2023evaluating}, HallB~\cite{liu2023hallusionbench}, MME~\cite{fu2023mme}, MathVista~\cite{lu2023mathvista}, MMB~\cite{liu2023mmbench}, MMB$^{\text{CN}}$~\cite{liu2023mmbench}, MM-Vet~\cite{yu2023mm}, and LLaVA$^{\text{W}}$~\cite{liu2023visual}. Note that, LLaVA$^{\text{W}}$ is newly evaluated with \textit{GPT-4-0613} because the original evaluator \textit{GPT-4-0314} is deprecated.}
\vspace{-5mm}
\label{tab:1}
\end{table*}

\section{\troltitle TroL: Traversal of Layers}
\label{sec:method}

\paragraph{Model Architecture.} As illustrated in Figure~\ref{fig:figure4}, \trol TroL is composed of a vision encoder, a vision projector, and a backbone multimodal large language model (MLLM) based on a pre-trained LLM. We utilize CLIP-L~\citep{clip} and InternViT~\citep{chen2023internvl} for the vision encoder, which are text-aligned vision encoders based on image-text contrastive learning with a small text encoder (CLIP) and QLLaMA-8B~\citep{cui2023efficient} (InternViT), respectively. For the vision projector, we employ two fully-connected layers with the GELU activation function~\citep{hendrycks2016gaussian}. As for the backbone multimodal LLM, we use Phi-3-mini~\citep{abdin2024phi} with a 3.8B model size, and InternLM2~\citep{2023internlm, cai2024internlm2} with 1.8B and 7B model sizes. 3.3T and 2T tokens are used during the pre-training of these LLMs, respectively.

\paragraph{Visual Instruction Tuning Dataset.} We gather a wide range of visual instruction tuning datasets requiring diverse capabilities such as fundamental image understanding, common-sense knowledge, non-object concepts (\textit{e.g.,} charts, diagrams, documents, signs, symbols), math problems, and their integrated capabilities. This is because we aim to make \trol TroL encompass diverse capabilities for vision language tasks despite its efficient model size. To balance the dataset samples across numerous capabilities, we selectively choose samples from existing visual instruction tuning datasets: ShareGPT4V-Caption/Instruct~\citep{chen2023sharegpt4v}, ALLaV4V-Text~\citep{chen2024allava}, MiniGemini-Instruct~\citep{li2024mini}, Doc-Downstream/Reason~\citep{hu2024mplug}, GLLaVA-Align/Instruct~\citep{gao2023g}, and Math-Vision/Instruct/Plus~\citep{wang2024measuring, yue2023mammoth, yue2024mammoth2}. In summary, we collect 899K real-world image/text-only samples, 627K samples for documents, charts, diagrams, signs, and symbols, and 747K math samples (180.5K with images and 566.8K text-only). Overall, the number of visual instruction tuning samples we used to build \trol TroL totals 2.3M samples.

\begin{table*}[t!]
\vspace{-6mm}
\centering
\resizebox{\linewidth}{!}{
\renewcommand{\tabcolsep}{0.8mm}
\begin{tabular}{lccccccccccccc}
\toprule
LLVMs     & Q-Bench & SQA$^{\text{I}}$ & AI2D & ChartQA & SEED$^{\text{I}}$ & POPE & HallB & MME & MathVista & MMB & MMB$^{\text{CN}}$ & MM-Vet & LLaVA$^{\text{W}}$  \\
\midrule
UIO-2-XL-3.2B~\cite{lu2023unifiedio}              & -    & 87.4 & -    & -    & 60.2 & 87.2 & -    & -    & -    & 68.1 & -    & -    & -       \\
Gemini Nano-2-3.2B~\cite{team2023gemini}          & -    & -    & -    & -    & -    & -    & -    & -    & 30.6 & -    & -    & -    & -       \\
MobileVLM-3B~\cite{chu2023mobilevlm}              & -    & 61.2 & -    & -    & -    & 84.9 & -    & -    & -    & 59.6 & -    & -    & -       \\
MobileVLM-V2-3B~\cite{chu2024mobilevlm}           & -    & 70.0 & -    & -    & -    & 84.7 & -    & -    & -    & 63.2 & -    & -    & -       \\
MoE-LLaVA-2.7B$\times$4~\cite{lin2024moe}         & -    & 70.3 & -    & -    & -    & 85.7 & -    & -    & -    & 68.0 & -    & 35.9 & -       \\
LLaVA-Phi-2.7B~\cite{zhu2024llava}                & -    & 68.4 & -    & -    & -    & 85.0 & -    & -    & -    & 59.8 & -    & 28.9 & -       \\
Imp-v1-3B~\cite{shao2024imp}                      & -    & 70.0 & -    & -    & -    & \textbf{88.0} & -    & -    & -    & 66.5 & -    & 33.1 & -       \\
TinyLLaVA-3.1B~\cite{zhou2024tinyllava}           & -    & 69.1 & -    & -    & -    & 86.4 & -    & -    & -    & 66.9 & -    & 32.0 & -       \\
TinyLLaVA-Sig-Phi-3.1B~\cite{zhou2024tinyllava}   & -    & 69.1 & -    & -    & -    & 86.4 & -    & -    & -    & 66.9 & -    & 32.0 & -       \\
Bunny-3B~\cite{he2024efficient}                   & -    & 70.9 & 38.2 & -    & 62.5 & 86.8 & -    & 1778 & -    & 68.6 & -    & -    & -       \\
MiniCPM-2.4B~\cite{hu2024minicpm}                 & -    & -    & 56.3 & -    & -    & -    & -    & 1650 & 28.9 & 64.1 & 62.6 & 31.1 & -       \\
MiniCPM-V2-2.8B~\cite{hu2024minicpm}              & -    & -    & 62.9 & -    & -    & -    & -    & 1809 & 38.7 & 69.1 & 66.5 & 41.0 & -       \\
MM1-3B~\cite{mckinzie2024mm1}                     & -    & 69.4 & -    & -    & 68.8 & 87.4 & -    & 1762 & 32.0 & 67.8 & -    & 43.7 & -       \\
MM1-MoE-3B$\times$64~\cite{mckinzie2024mm1}       & -    & 76.1 & -    & -    & 69.4 & 87.6 & -    & 1773 & 32.6 & 70.8 & -    & 42.2 & -       \\
ALLaVA-3B~\cite{chen2024allava}                   & -    & -    & -    & -    & 65.2 & -    & -    & 1623 & -    & 64.0 & -    & 32.2 & -       \\
ALLaVA-3B-Longer~\cite{chen2024allava}            & -    & -    & -    & -    & 65.6 & -    & -    & 1564 & -    & 64.6 & -    & 35.5 & -       \\
\midrule
\rowcolor{Green}
TroL-3.8B   & \textbf{70.0} 
            & \textbf{90.8}
            & \textbf{73.6}    
            & \textbf{73.8}
            & \textbf{70.5} 
            & 86.5
            & \textbf{62.2}    
            & \textbf{1980}
            & \textbf{55.1}
            & \textbf{79.2} 
            & \textbf{77.1}  
            & \textbf{51.1} 
            & \textbf{76.6}    \\
\midrule
UIO-2-L-1.1B~\cite{lu2023unifiedio}               & -    & 78.6 & -    & -    & 51.1 & 77.8 & -    & -    & -    & 62.1 & -    & -    & -        \\
MobileVLM-1.7B~\cite{chu2023mobilevlm}            & -    & 57.3 & -    & -    & -    & 84.5 & -    & -    & -    & 53.2 & -    & -    & -        \\
MobileVLM-V2-1.7B~\cite{chu2024mobilevlm}         & -    & 66.7 & -    & -    & -    & 84.3 & -    & -    & -    & 57.7 & -    & -    & -        \\
DeepSeek-VL-1.3B~\cite{lu2024deepseek}            & -    & -    & -    & -    & 66.7 & 87.6 & -    & -    & 31.1 & 64.6 & 62.9 & 34.8 & -        \\
Mini-Gemini-2B~\cite{li2024mini}                  & -    & -    & -    & -    & -    & -    & -    & 1653 & 29.4 & 59.8 & -    & -    & -       \\
MoE-LLaVA-1.8B$\times$4~\cite{lin2024moe}         & -    & 63.1 & -    & -    & -    & 87.0 & -    & -    & -    & 59.7 & -    & 25.3 & -       \\
\midrule
\rowcolor{Green}
TroL-1.8B   & \textbf{68.2} 
            & \textbf{87.5}
            & \textbf{68.9}    
            & \textbf{64.0}
            & \textbf{69.0} 
            & \textbf{88.6}
            & \textbf{60.1}    
            & \textbf{2038}
            & \textbf{45.4}
            & \textbf{76.1} 
            & \textbf{74.1}  
            & \textbf{45.1} 
            & \textbf{69.7}    \\
\bottomrule

\end{tabular}
}
\vspace{-2mm}
\caption{Comparison with the current existing smaller open-source LLVMs across 1B$\sim$4B model sizes, evaluating vision language performances of \trol TroL on numerous evaluation benchmarks equally used in Table~\ref{tab:1}. Note that, the compared baselines were not validated on Q-Bench, ChartQA, HallB, and LLaVA$^{\text{W}}$, but we measure their performances to compare with the baselines mentioned in Table~\ref{tab:1}.}
\vspace{-5mm}
\label{tab:2}
\end{table*}
\begin{figure}[t!]
\vspace{0mm}
    \centering
    \includegraphics[width=0.4\textwidth]{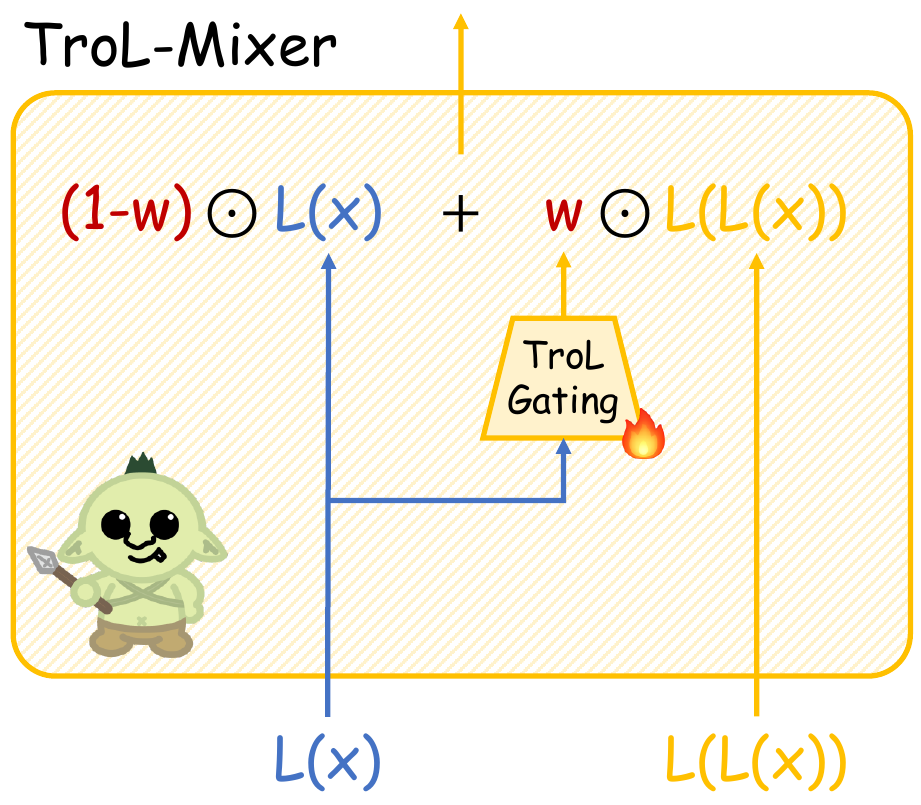}
    \vspace{-2mm}
    \caption{Overview of mix operation in TroL-Mixer}
    \label{fig:figure5}
    \vspace{-5mm}
\end{figure}

\paragraph{Layer Traversing.} To effectively enlarge learning capabilities with smaller sized LLVMs, we introduce a layer traversing technique that allows the reuse of layers. As described in Figure~\ref{fig:figure1}, once the input token $x\in\mathbb{R}^{N\times D}$ (\textit{i.e.,} vision language features), where $N$ denotes the number of tokens and $D$ denotes the hidden dimension of a layer, is given, then a layer normally outputs $L(x)$ from the input token. Layer traversing makes the output of the layer forward in the equal layer once again: $L(L(x))$. Subsequently, the outputs $L(x)$ and $L(L(x))$ from the equal layer get mixed to further improve the vision language features by themselves. Here, we present TroL-Mixer that provides a mixing operation with $L(x)$ and $L(L(x))$, where TroL Gating is introduced to determine how much reused vision language feature $L(L(x))$ is needed for the next layer, by looking at the feature status of the first propagation output $L(x)$. More specifically, the output $L(x)$ is propagated into the TroL Gating and it produces a mixing ratio $w\in\mathbb{R}^{N}$ for each token. It is used for token-wise multiplication $\odot$ with $L(x)$ and $L(L(x))$. Finally, the mixed output $(1-w) \odot L(x) + w\odot L(L(x))$ in Figure~\ref{fig:figure5} is used to propagate to the next layer and it is continually operated as it goes layers. By doing this, we expect the layer traversing to stimulate the effect of retracing and looking back at the answering stream once again, thereby enlarging the learning capabilities by nature.

\paragraph{Training Strategy.} TroL-Layer is applied to backbone multimodal LLMs described in Figure~\ref{fig:figure1} and \ref{fig:figure4} for application to layer traversing technique. Thereafter, we conduct a two-step training process to effectively implement layer traversing using LLVMs, creating a new efficient LLVM family named \trol TroL. In the first training step, we focus on training a vision projector and all TroL-Mixers for every TroL-Layer. This step is essential as it aligns vision and language information while synchronizing the TroL-Mixers with the response stream in the backbone multimodal LLMs, thus facilitating the understanding of layer traversing operation. The subsequent second training step involves additional training of these elements alongside the backbone multimodal LLMs together.

\section{Experiment}
\label{sec:experi}

\begin{table*}[t!]
\vspace{-6mm}
\centering
\begin{minipage}[t]{\linewidth}

\begin{minipage}[t]{0.99\linewidth}
\newcolumntype{g}{>{\columncolor{Green}}c}
\resizebox{\linewidth}{!}{
\renewcommand{\tabcolsep}{1mm}
\begin{tabular}{lcccccccggg}
\toprule
Benchmarks & OmniFusion-7B & DeepSeek-VL-7B & MoVA-7B & ASMv2-7B & LAF-7B & CoLLaVO-7B & MoAI-7B & TroL-1.8B & TroL-3.8B & TroL-7B \\
\midrule
POPE       & 87.2      & 88.1  & 88.6  & 86.3  & \textbf{88.8}   & 87.2    & 87.1 & 88.6 & 86.5 & 87.8\\
\cdashline{1-9}\noalign{\vskip 0.5ex}
SQA-IMG    & 69.2      & 57.7  & 74.4  & 87.1  & -      & 80.7    & 83.5 & 87.5 & 90.8 & \textbf{92.8}\\
\cdashline{1-9}\noalign{\vskip 0.5ex}
LLaVA-W    & -         & -     & -     & 78.9  & -      & 69.5    & 71.9 & 69.7 & 76.6 & \textbf{92.8}\\
\cdashline{1-9}\noalign{\vskip 0.5ex}
MM-Vet     & 39.4      & 41.5  & -     & 41.3  & 38.9   & 40.3    & 43.7 & 45.1 & 51.1 & \textbf{54.7}\\
\cdashline{1-9}\noalign{\vskip 0.5ex}
MMStar     & -         &  -    & -     & -     & -      & 42.1    & 48.7 & 45.5 & 46.5 & \textbf{51.3}\\
\bottomrule
\end{tabular}
}
\vspace{-3mm}
\caption*{(a) Comparison with LLVMs using additional modules: OmniFusion~\cite{goncharova2024omnifusion}, DeepSeek-VL~\cite{lu2024deepseek}, MoVA~\cite{kar2024brave}, ASMv2~\cite{wang2024all}, LAF~\cite{jiao2024enhancing}, CoLLaVO~\cite{lee2024collavo}, and MoAI~\cite{lee2024moai}}
\end{minipage}

\begin{minipage}[t]{0.99\linewidth}
\resizebox{\linewidth}{!}{
\renewcommand{\tabcolsep}{3mm}
\begin{tabular}{lccccccc}
\toprule
LLVMs      & Recognition & OCR  & Knowledge & Language Generation & Spatial Awareness & Math Problems & Avg \\
\midrule
CoLLaVO-7B~\cite{lee2024collavo} & 45.6        & 31.1 & 29.8      & 30.2                & 37.9              & 5.8  & 41.0 \\
\cdashline{1-8}\noalign{\vskip 0.5ex}
MoAI-7B~\cite{lee2024moai}    & 48.3        & 34.8 & 33.5      & 33.0                & 39.7              & 7.7  & 43.7 \\
\midrule
\rowcolor{Green}
TroL-1.8B  & 42.0
         & 48.2
         & 31.9
         & 29.0
         & 47.1
         & 41.2
         & 45.1 \\
         \rowcolor{Green}
\rowcolor{Green}
TroL-3.8B& 45.7
         & \textbf{56.6}
         & 37.0
         & 40.6
         & \textbf{56.5}
         & 48.1
         & 51.1 \\
\rowcolor{Green}
TroL-7B  & \textbf{54.2}
         & 54.6
         & \textbf{42.4}
         & \textbf{49.3}
         & 52.7
         & \textbf{53.8}
         & \textbf{54.7} \\
\bottomrule
\end{tabular}
}
\vspace{-3mm}
\caption*{(b) Evaluating sub-benchmark in MM-Vet~\cite{yu2023mm} with LLVMs utilizing computer vision models}
\end{minipage}

{\begin{minipage}[t]{0.49\linewidth}
\resizebox{\linewidth}{!}{
\renewcommand{\tabcolsep}{1mm}
\begin{tabular}{lccccccc}
\toprule
LLVMs            & CP                       & FP                       & IR                       & LR                       & ST                       & MA                       & Avg                      \\
\midrule
Yi-VL-34B~\cite{young2024yi}        & \multicolumn{1}{l}{53.2} & \multicolumn{1}{l}{31.2} & \multicolumn{1}{l}{52.0} & \multicolumn{1}{l}{32.4} & \multicolumn{1}{l}{12.4} & \multicolumn{1}{l}{35.2} & \multicolumn{1}{l}{36.1} \\
\cdashline{1-8}\noalign{\vskip 0.5ex}
CogVLM-Chat-17B~\cite{wang2023cogvlm}  & \multicolumn{1}{l}{66.8} & \multicolumn{1}{l}{36.8} & \multicolumn{1}{l}{49.2} & \multicolumn{1}{l}{31.2} & \multicolumn{1}{l}{23.6} & \multicolumn{1}{l}{11.6} & \multicolumn{1}{l}{36.5} \\
\cdashline{1-8}\noalign{\vskip 0.5ex}
SPHINX-MoE-7B$\times$8~\cite{gao2024sphinx} & 58.4                     & 40.8                     & 47.6                     & 35.2                     & 19.2                     & 32.0                     & 38.9                     \\
\cdashline{1-8}\noalign{\vskip 0.5ex}
InternVL1.2-40B~\cite{chen2023internvl}  & 67.6                     & 43.2                     & 61.2                     & \textbf{47.2}                     & 24.0                     & 19.2                     & 43.7                     \\
\cdashline{1-8}\noalign{\vskip 0.5ex}
LLaVA-NeXT-34B~\cite{liu2024llavanext}   & 66.4                     & \textbf{52.0}            & 62.4                     & 46.0                     & 32.4                     & \textbf{53.6}            & \textbf{52.1}                     \\
\midrule
\rowcolor{Green}
TroL-1.8B        & 63.2
               & 41.6
               & 59.2
               & \textbf{47.2}
               & 30.0
               & 31.6
               & 45.5        \\
\rowcolor{Green}
TroL-3.8B        & 61.2
               & 40.0
               & 54.0
               & 43.2
               & 31.2
               & 49.6
               & 46.5        \\
\rowcolor{Green}
TroL-7B        & \textbf{69.2}
               & 47.6
               & \textbf{65.6}
               & 45.2
               & \textbf{37.2}
               & 42.8
               & 51.3        \\
\bottomrule
\end{tabular}
}
\vspace{-3mm}
\caption*{(c) MMStar~\cite{chen2024we}}
\end{minipage}
\begin{minipage}[t]{0.49\linewidth}
\resizebox{\linewidth}{!}{
\renewcommand{\tabcolsep}{1mm}
\begin{tabular}{lccccccc}
\toprule
LLVMs          & TD   & TL   & TO   & VI   & VD   & VO   & Avg  \\
\midrule
G-LLaVA-7B~\cite{gao2023g}                 & 20.9 & 20.7 & 21.1 & 17.2 & 16.4 & 9.4  & 16.6 \\
\cdashline{1-8}\noalign{\vskip 0.5ex}
LLaVA-NeXT-13B~\cite{liu2024llavanext}     & 12.8 & 12.0 & 9.9  & 10.7 & 9.7  & 6.3  & 10.3 \\
\cdashline{1-8}\noalign{\vskip 0.5ex}
ShareGPT4V-13B~\cite{chen2023sharegpt4v}   & 16.2 & 16.2 & 6.6  & 15.5 & 13.8 & 3.7  & 13.1 \\
\cdashline{1-8}\noalign{\vskip 0.5ex}
SPHINX-Plus-13B~\cite{gao2024sphinx}        & 13.9 & 11.6 & 14.9 & 11.6 & 13.5 & 10.4 & 12.2 \\
\cdashline{1-8}\noalign{\vskip 0.5ex}
SPHINX-MoE-7B$\times$8~\cite{gao2024sphinx}        & 26.2 & 17.4 & 26.7 & 16.7 & 12.5 & 11.1 & 16.8 \\
\midrule
\rowcolor{Green}
TroL-1.8B                                & 26.1
                                         & 26.5
                                         & 25.5
                                         & 25.6
                                         & 25.6
                                         & 14.8
                                         & 24.0 \\
\rowcolor{Green}
TroL-3.8B                                & \textbf{42.3}
                                         & \textbf{38.8}
                                         & \textbf{40.6}
                                         & \textbf{35.5}
                                         & \textbf{35.9}
                                         & \textbf{21.4}
                                         & \textbf{35.8} \\
\rowcolor{Green}
TroL-7B                                  & 37.8
                                         & 34.1
                                         & 36.9
                                         & 32.1
                                         & 32.1
                                         & 19.5
                                         & 32.1 \\
\bottomrule
\end{tabular}
}
\vspace{-3mm}
\caption*{(d) MathVerse~\cite{zhang2024mathverse}}
\end{minipage}
}

\begin{minipage}[t]{0.99\linewidth}
\resizebox{\linewidth}{!}{
\renewcommand{\tabcolsep}{5mm}
\begin{tabular}{lcccccccc}
\toprule
\multirow{2}{*}{LLVMs} & \multicolumn{3}{c}{Website} & \multicolumn{2}{c}{Element} & \multicolumn{2}{c}{Action} & \multirow{2}{*}{Average} \\
\cmidrule(lr){2-4}\cmidrule(lr){5-6}\cmidrule(lr){7-8}
                                           & Caption  & WebQA & HeadOCR & OCR         & Ground        & Prediction     & Ground    &                          \\
\midrule
InstrcutBLIP-13B~\cite{dai2023instructblip}& 11.6     & 5.2   & 7.6     & 6.0         & 11.4          & 11.4           & 17.5      & 10.1                     \\
Yi-VL-6B~\cite{young2024yi}                & 8.0      & 14.3  & 43.8    & 3.5         & 16.2          & 13.9           & 13.6      & 16.2                     \\
\rowcolor{Gray}
LLaVA1.5-7B~\cite{liu2023improved}         & 15.3     & 13.2  & 41.0    & 5.7         & 12.1          & 17.8           & 13.6      & 17.0                     \\
\rowcolor{Gray}
LLaVA1.5-13B ~\cite{liu2023improved}       & 20.0     & 16.2  & 41.1    & 11.8        & 15.0          & 22.8           & 8.7       & 19.4                     \\
CogVLM-17B~\cite{wang2023cogvlm}           & 16.6     & 30.6  & 65.9    & 10.0        & 17.7          & 11.7           & 23.3      & 25.1                     \\
VILA-13B~\cite{lin2023vila}                & 12.7     & 28.8  & 67.9    & 12.6        & 16.5          & 36.3           & 16.5      & 27.3                     \\
DeepSeek-VL-7B~\cite{lu2024deepseek}       & 18.1     & 30.0  & 63.4    & 18.1        & 16.2          & 35.2           & 15.5      & 28.1                     \\
\rowcolor{Gray}
LLaVA-NeXT-7B~\cite{liu2024llavanext}      & \textbf{27.0}     & 39.8  & 57.3    & 54.8        & 31.7          & 30.6           & 10.7      & 36.0                     \\
\rowcolor{Gray}
LLaVA-NeXT-13B~\cite{liu2024llavanext}     & 26.5     & 44.5  & 52.8    & 56.1        & 31.7          & 48.4           & 15.5      & 39.4                     \\
\rowcolor{Gray}
LLaVA-NeXT-34B~\cite{liu2024llavanext}     & 24.3     & 48.2  & 67.1    & \textbf{71.9}        & 43.1          & \textbf{74.0}           & 25.2      & \textbf{50.5}                     \\
\midrule
\rowcolor{Green}
TroL-1.8B              & 14.2   
                       & 38.3
                       & 58.5   
                       & 29.5
                       & 24.7
                       & 14.2
                       & 29.1
                       & 29.8       \\
\rowcolor{Green}
TroL-3.8B              & 22.5    
                       & \textbf{65.3}
                       & 70.2   
                       & 63.0
                       & \textbf{69.7}
                       & 20.3
                       & \textbf{39.8}
                       & 50.1 \\
\rowcolor{Green}
TroL-7B                & 23.6    
                       & 44.7 
                       & \textbf{74.4}   
                       & 38.6
                       & 40.9
                       & 16.0
                       & 32.3
                       & 38.6       \\
\bottomrule
\end{tabular}
}
\vspace{-3mm}
\caption*{(e) Evaluating sub-benchmark in VisualWebBench~\cite{liu2024visualwebbench} with numerous open-source LLVMs.}
\end{minipage}

\end{minipage}
\vspace{-4mm}
\caption{Detailed comparison of \trol TroL across challenging evaluation benchmarks. Note that, the sub-benchmark category names in (c) and (d) are represented in Appendix~\ref{app:A}. Note that, Phi-3-mini~\cite{abdin2024phi} built in TroL-3.8B has shown excellent benefits for understanding web pages and solving human-level math problems.}
\label{tab:4}
\vspace{-3mm}
\end{table*}

\paragraph{Implementation Detail.} To ensure successful reproducibility, we present four key technical aspects of \trol TroL: the detailed structure of (a) backbone multimodal LLMs, (b) vision encoder, vision projectors, TroL Gating. In addition, the detailed procedures of (c) training and inference are described. 

\paragraph{(a)} For backbone multimodal LLMs, we employ Phi-3-mini~\citep{abdin2024phi} and InternLM2~\citep{2023internlm, cai2024internlm2}, where Phi-3-mini 3.8B model consists of 32 layers with hidden dimension of 3072, while InternLM2-1.8B | 7B features 24 | 32 layers with hidden dimension of 2048 | 4096, respectively.

\paragraph{(b)} We use CLIP-L~\citep{clip} and InternViT~\citep{chen2023internvl} as vision encoders, each comprising 428M | 300M parameters, with 24 layers and hidden dimension of 1024. When investigating best structural combination, we consider CLIP-L for 1.8B and 7B model and consider InternViT for 3.8B model. The vision projector consists of an MLP that adjusts the hidden dimension from 1024 to 2048 | 3072 | 4096 to match the hidden dimension of backbone multimodal LLMs. This MLP contains two fully-connected layers with GELU activation~\citep{hendrycks2016gaussian}. TroL Gating employs a single fully-connected layer that converts the hidden dimension from 2048 | 3072 | 4096 to 1, resulting in a total of $2048\times 24=$49K, $3072\times 32=$98K, and $4096\times 32=$131K parameters for TroL Gating each, which are minimal compared to the 1.8B, 3.8B, and 7B model sizes.

\paragraph{(c)} The training and evaluation of \trol TroL are conducted in a computing environment with 8$\times$NVIDIA Tesla A100 80GB and 8$\times$NVIDIA RTX A6000 48GB, each. To optimize the training process, we use one epoch of training for each step under 4/8-bit quantization and bfloat16 data type~\citep{kalamkar2019study} for each backbone multimodal LLM: \trol TroL-1.8B (4-bit), \trol TroL-3.8B (8-bit), and \trol TroL-7B (4-bit). The 4-bit quantization employs double quantization and normalized float 4-bit (nf4)\citep{dettmers2023qlora}. Additionally, QLoRA~\citep{hu2021lora, dettmers2023qlora} is used to train the multimodal LLMs with 64 rank and 64 alpha parameters. We apply the AdamW optimizer~\citep{loshchilov2018decoupled} and use cosine annealing to schedule the learning rate from 1e-4 to 1e-6 in each training step. We also utilize gradient checkpointing~\citep{sohoni2019low} for efficient memory usage. With a gradient accumulation of 6, batch sizes are totally set to 768 for each training step, and each step takes approximately one to three days according to the model sizes. For efficient inference, we validate \trol TroL under the same quantization bit during training, and we employ deterministic beam search~\citep{freitag-al-onaizan-2017-beam} ($n=5$) for text generation. Moreover, in inference, we apply layer traversing technique only to the user questions in order to avoid dramatically increased inference time. Note that, \trol TroL is implemented on the efficient propagation, FlashAttention2~\citep{dao2022flashattention, dao2023flashattention} for speed-up attention computation.

\begin{table*}[t!]
\vspace{-6mm}
\centering
\begin{minipage}[t]{0.31\linewidth}
\resizebox{\linewidth}{!}{
\renewcommand{\tabcolsep}{2.8mm}
\begin{tabular}{lrcc}
\toprule
LLMs       & Param & MMStar& MM-Vet             \\
\midrule
Gemma      & 2B    & 44.6  & 42.8             \\
\rowcolor{Green}
InternLM2  & 1.8B  & \textbf{45.5}  & \textbf{45.1}    \\
Qwen1.5    & 4B    & 45.5  & 49.9             \\
\rowcolor{Green}
Phi-3-mini & 3.8B  & \textbf{46.5}  & \textbf{51.1}    \\
Mistral    & 7B    & 49.7  & 53.2             \\
\rowcolor{Green}
InternLM2  & 7B    & \textbf{51.3}  & \textbf{54.7}    \\
\bottomrule
\end{tabular}
}
\vspace{-3mm}
\caption*{(a) Backbone pre-trained LLMs}

\vspace{2mm}
\resizebox{\linewidth}{!}{
\renewcommand{\tabcolsep}{3.5mm}
\begin{tabular}{lcc}
\toprule
Operation                         & MMStar& MM-Vet \\
\midrule
$(1-w)\odot L(x)$                 & 40.8  & 45.4   \\
$w \odot L(L(x))$                 & 44.3  & 48.1   \\
Random $w$                        & 40.3  & 37.0   \\
Uniform $w$                       & 41.4  & 46.3   \\
Learnable $w$                     & 46.7  & 49.9   \\
\rowcolor{Green}
Figure~\ref{fig:figure5}          & \textbf{51.3} & \textbf{54.7}    \\
\bottomrule
\end{tabular}
}
\vspace{-3mm}
\caption*{(d) Mixing operation}

\end{minipage}
\begin{minipage}[t]{0.31\linewidth}
\resizebox{\linewidth}{!}{
\renewcommand{\tabcolsep}{3.3mm}
\begin{tabular}{lccc}
\toprule
Family    & Lay-Trav& MMStar& MM-Vet \\
\midrule
TroL-1.8B & \xmark  & 36.0         & 34.6   \\
\rowcolor{Green}
TroL-1.8B & \cmark  & \textbf{45.5}& \textbf{45.1}   \\
\midrule
TroL-3.8B & \xmark  & 37.4         & 43.5   \\
\rowcolor{Green}
TroL-3.8B & \cmark  & \textbf{46.5}& \textbf{51.1}   \\
\midrule
TroL-7B   & \xmark  & 41.2         & 45.8   \\
\rowcolor{Green}
TroL-7B   & \cmark  & \textbf{51.3} & \textbf{54.7}   \\
\bottomrule
\end{tabular}
}
\vspace{-3mm}
\caption*{(b) Use of layer traversing}

\vspace{2mm}
\resizebox{\linewidth}{!}{
\renewcommand{\tabcolsep}{5.9mm}
\begin{tabular}{lcc}
\toprule
Percent & MMStar & MM-Vet \\
\midrule
10\%   & 39.9   & 41.5   \\
30\%   & 46.3   & 48.2   \\
50\%   & 50.8   & 54.0   \\
70\%   & 51.2   & 54.4   \\
90\%   & \textbf{51.3}   & 54.6   \\
\rowcolor{Green}
100\%  & \textbf{51.3}   & \textbf{54.7}    \\
\bottomrule
\end{tabular}
}
\vspace{-3mm}
\caption*{(e) Training percentage}

\end{minipage}
\begin{minipage}[t]{0.31\linewidth}
\resizebox{\linewidth}{!}{
\renewcommand{\tabcolsep}{2mm}
\begin{tabular}{lrcc}
\toprule
Structure                  & Param & MMStar& MM-Vet \\
\midrule
PR $\times$2               & 13B   & \textbf{52.3}& \textbf{57.4}   \\
PR                         & 6B    & 52.2& 57.1   \\
MHA$\times$2 + FC          & 5B    & 51.8& 56.8   \\
MHA + FC                   & 2B    & 51.7& 56.6   \\
FC$\times$2                & 537M  & 51.3& 54.8   \\
\rowcolor{Green}
FC                         & 131K  & 51.3& 54.7    \\
\bottomrule
\end{tabular}
}
\vspace{-3mm}
\caption*{(c) TroL Gating}

\vspace{2mm}
\centering
\resizebox{\linewidth}{!}{
\renewcommand{\tabcolsep}{1.0mm}
\begin{tabular}{llc}
\toprule
LLVMs                      & VRAM          & Time Ratio\\
\midrule
\rowcolor{Gray}
InternLM2-1.8B             & 6GB           & 1.0   \\
+CLIP-L (Image Tokens)     & +1GB          & 1.1   \\
\rowcolor{Green}
+Layer Traversing          & +1GB          & 1.3   \\
\midrule
\rowcolor{Gray}
Phi-3-mini-3.8B            & 7GB           & 1.0   \\
+InternViT (Image Tokens)  & +1GB          & 1.2   \\
\rowcolor{Green}
+Layer Traversing          & +1GB          & 1.4   \\
\bottomrule
\end{tabular}
}
\vspace{-3mm}
\caption*{(f) Inference speed}

\end{minipage}
\vspace{-2mm}
\caption{Ablation studies to identify the effectiveness of \trol TroL by controlling the six main factors.}
\label{tab:5}
\vspace{-3mm}
\end{table*}

\begin{figure*}[t!]
\vspace{0mm}
    \centering
    \includegraphics[width=\textwidth]{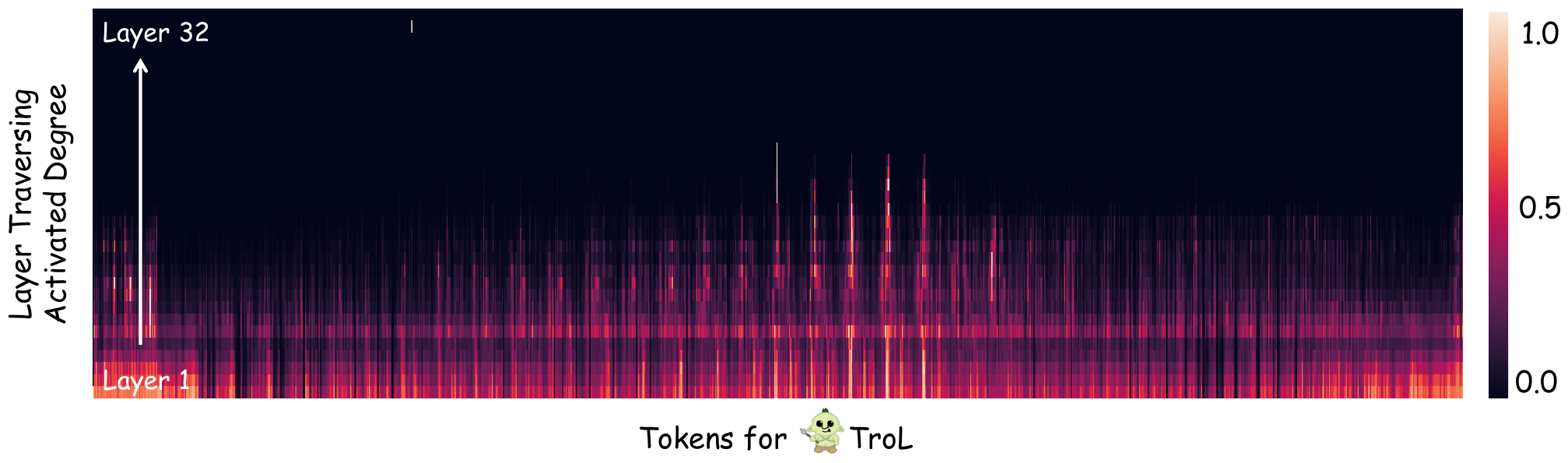}
    \vspace{-7mm}
    \caption{Visualization of layer traversing activated degree in each layer where mixing ratios of $w$ in token-wise manner are shown as the colors in color bar. To clearly discriminate where the layer traversing mostly occurs, they are applied to min-max normalization in each layer. (\textit{i.e.,} The brighter they are, the more layer traversing happens)}
    \label{fig:figure6}
    \vspace{-5mm}
\end{figure*}

\paragraph{Validation on Evaluation Benchmarks.} We have demonstrated  super vision language performances of \trol TroL, despite its limited model size, as depicted in Table~\ref{tab:1}, Table~\ref{tab:2}, and Table~\ref{tab:4}. This is attributed to the enhanced learning capabilities by layer traversing. Furthermore, Table~\ref{tab:5} have shown several ablation studies to clearly corroborate the effectiveness of \trol TroL in light of seven factors: (a) backbone pre-trained LLMs, (b) the use of layer traversing, (c) the structure of TroL Gating, (d) mixing operation of TroL-Mixer, (e) training percentage of visual instruction tuning, and (f) ratio of measured time for inference speed. Note that, in Table~\ref{tab:5}(c), `FC' denotes a fully-connected layer, and `MHA' denotes a multi-head attention block, and `PR' denotes the structure of Perceiver Resampler~\citep{alayrac2022flamingo}. Through the ablation, we consider the efficiency of the model structure and performance on vision language tasks to build the current architectures of \trol TroL, including the pre-trained LLM, TroL Gating architecture, and the TroL-Mixer operation. Additionally, layer traversing focusing on user's questions during inference makes text generation speed comparable to the baselines we used for pre-trained LLM. Note that, all descriptions of the evaluation benchmarks used in this paper are explained in Appendix~\ref{app:A}, and we show diverse samples for \trol TroL's text generation quality in Appendix~\ref{app:B}. In addition, Appendix~\ref{app:C} provides further ablation studies to check more various factors.

\section{Discussion and Conclusion}
\label{sec:discussion_and_concludion}
A new LLVM family, \trol TroL, has demonstrated significant advancements in vision language performance despite its inherent limitation of having smaller layers compared to larger model sizes in both open- and closed-source LLVMs. Table~\ref{tab:4} suggests that reusing layers through layer traversing can be an effective alternative to incorporating additional modules. We expect \trol TroL to remain an efficient option in the rapidly evolving field, solidifying its place in the landscape of efficient LLVMs.

Interestingly, when analyzing all mixing ratios for each layer, we observed in Figure~\ref{fig:figure6} that the layer traversing event of looking back and retracing the answering stream mostly occurs in shallower layers, while the deeper layers are not involved in traversing. This suggests that recursively enhancing vision language features continues until they are fully matured; once matured, no need for anymore. Further, we plan to explore further methods in the shallower and deeper layers, which significantly enhance learning capabilities by virtually increasing the hidden dimension without physically enlarging it. We believe this approach, combined with layer traversing, could be one of the crucial keys serving as efficient LLVMs, potentially propelling \trol TroL to the forefront within 1$\sim$3B.

\clearpage

\section*{Acknowledgments}
This work was partially supported by two funds: Center for Applied Research in Artificial Intelligence (CARAI) grant funded by DAPA and ADD (UD230017TD) and IITP grant funded by the Korea government (MSIT) (RS-2022-II220984). Additionally, it was supported by the KISTI National Supercomputing Center with supercomputing resources including technical support (KSC-2024-CRE-0160).

\section{Limitations}
\label{sec:d and l}
\trol TroL, like several other LLVMs, naturally faces the challenge of the training computational burden. For building \trol TroL, we used high-end GPUs (8$\times$NVIDIA Tesla A100 80GB) and take at most three days in training despite using them. Practically, this challenge might be mitigated using numerous optimization tools and techniques~\citep{xue2024powerinfer2}, such as PagedAttention~\citep{kwon2023efficient}, ChunkAttention~\citep{ye2024chunkattention}, optimized CUDA kernel \& HIP graph, GPTQ~\citep{frantar2022gptq}, AWQ~\citep{lin2023awq}, SqueezeLLM~\citep{kim2023squeezellm}, dynamic KV cache, FP8 KV cache, and prefix caching. However, these techniques mentioned tend to be more applicable during the inference phase. Therefore, advanced techniques beyond quantization for reducing the training computational burden should be further developed to enable the AI community to handle large models more effectively. Beyond that, we expect \trol TroL to be equipped with numerous bootstrapping methods~\citep{lee2020training, lee2020towards,  kim2021distilling, lee2022masking, kim2023demystifying, lee2023mitigating, kim2023causal, kim2023mitigating, lee2024meteor, park2024robust, park2024integrating, kim2024improving}, providing a wide range of variations for both general and specific tasks.
\section{Ethics Statement}
We affirm that all research presented in this paper adheres to the highest principles of ethical conduct and integrity. The experiments conducted and the results reported are based on rigorous scientific methodologies, with the goal of contributing positively and responsibly to the field of large language and vision models (LLVMs). All datasets utilized in this study, including visual instruction tuning datasets~\citep{chen2023sharegpt4v, chen2024allava, li2024mini, hu2024mplug, gao2023g, wang2024measuring, yue2023mammoth, yue2024mammoth2}, were acquired and analyzed in strict compliance with relevant regulations and guidelines governing research ethics and data privacy. We ensured that all data handling processes protected the confidentiality and rights of individuals represented in the datasets. Furthermore, we have transparently discussed any potential limitations and ethical considerations in Section~\ref{sec:d and l}, \textit{Limitations}. This commitment to transparency allows for a critical assessment of our work and underscores our dedication to maintaining the highest standards of integrity and accountability. We are committed to respecting and considering the impact of our research on communities and individuals. Our goal is to advance the field responsibly, with a conscientious approach that values ethical considerations as much as scientific innovation. By upholding these principles, we aim to foster trust and integrity within AI community.
\clearpage
\bibliography{ref}

\begin{thebibliography}{119}
\providecommand{\natexlab}[1]{#1}

\bibitem[{Abdin et~al.(2024)Abdin, Jacobs, Awan, Aneja, Awadallah, Awadalla, Bach, Bahree, Bakhtiari, Behl et~al.}]{abdin2024phi}
Marah Abdin, Sam~Ade Jacobs, Ammar~Ahmad Awan, Jyoti Aneja, Ahmed Awadallah, Hany Awadalla, Nguyen Bach, Amit Bahree, Arash Bakhtiari, Harkirat Behl, et~al. 2024.
\newblock Phi-3 technical report: A highly capable language model locally on your phone.
\newblock \emph{arXiv preprint arXiv:2404.14219}.

\bibitem[{Achiam et~al.(2023)Achiam, Adler, Agarwal, Ahmad, Akkaya, Aleman, Almeida, Altenschmidt, Altman, Anadkat et~al.}]{achiam2023gpt}
Josh Achiam, Steven Adler, Sandhini Agarwal, Lama Ahmad, Ilge Akkaya, Florencia~Leoni Aleman, Diogo Almeida, Janko Altenschmidt, Sam Altman, Shyamal Anadkat, et~al. 2023.
\newblock Gpt-4 technical report.
\newblock \emph{arXiv preprint arXiv:2303.08774}.

\bibitem[{Alayrac et~al.(2022)Alayrac, Donahue, Luc, Miech, Barr, Hasson, Lenc, Mensch, Millican, Reynolds et~al.}]{alayrac2022flamingo}
Jean-Baptiste Alayrac, Jeff Donahue, Pauline Luc, Antoine Miech, Iain Barr, Yana Hasson, Karel Lenc, Arthur Mensch, Katherine Millican, Malcolm Reynolds, et~al. 2022.
\newblock Flamingo: a visual language model for few-shot learning.
\newblock \emph{Advances in Neural Information Processing Systems}, 35:23716--23736.

\bibitem[{Bai et~al.(2023)Bai, Bai, Yang, Wang, Tan, Wang, Lin, Zhou, and Zhou}]{bai2023qwen}
Jinze Bai, Shuai Bai, Shusheng Yang, Shijie Wang, Sinan Tan, Peng Wang, Junyang Lin, Chang Zhou, and Jingren Zhou. 2023.
\newblock Qwen-vl: A frontier large vision-language model with versatile abilities.
\newblock \emph{arXiv preprint arXiv:2308.12966}.

\bibitem[{Cai et~al.(2024)Cai, Cao, Chen, Chen, Chen, Chen, Chen, Chen, Chen, Chu et~al.}]{cai2024internlm2}
Zheng Cai, Maosong Cao, Haojiong Chen, Kai Chen, Keyu Chen, Xin Chen, Xun Chen, Zehui Chen, Zhi Chen, Pei Chu, et~al. 2024.
\newblock Internlm2 technical report.
\newblock \emph{arXiv preprint arXiv:2403.17297}.

\bibitem[{Cao et~al.(2023)Cao, Paranjape, and Hajishirzi}]{cao2023pumer}
Qingqing Cao, Bhargavi Paranjape, and Hannaneh Hajishirzi. 2023.
\newblock \href {https://arxiv.org/abs/2305.17530} {Pumer: Pruning and merging tokens for efficient vision language models}.
\newblock \emph{Preprint}, arXiv:2305.17530.

\bibitem[{Chen et~al.(2024{\natexlab{a}})Chen, Xu, Kirmani, Ichter, Driess, Florence, Sadigh, Guibas, and Xia}]{chen2024spatialvlm}
Boyuan Chen, Zhuo Xu, Sean Kirmani, Brian Ichter, Danny Driess, Pete Florence, Dorsa Sadigh, Leonidas Guibas, and Fei Xia. 2024{\natexlab{a}}.
\newblock Spatialvlm: Endowing vision-language models with spatial reasoning capabilities.
\newblock \emph{arXiv preprint arXiv:2401.12168}.

\bibitem[{Chen et~al.(2024{\natexlab{b}})Chen, Chen, Zhang, Chen, Wu, Zhang, Chen, Li, Wan, and Wang}]{chen2024allava}
Guiming~Hardy Chen, Shunian Chen, Ruifei Zhang, Junying Chen, Xiangbo Wu, Zhiyi Zhang, Zhihong Chen, Jianquan Li, Xiang Wan, and Benyou Wang. 2024{\natexlab{b}}.
\newblock Allava: Harnessing gpt4v-synthesized data for a lite vision-language model.
\newblock \emph{arXiv preprint arXiv:2402.11684}.

\bibitem[{Chen et~al.(2023{\natexlab{a}})Chen, Zhang, Zeng, Zhang, Zhu, and Zhao}]{chen2023shikra}
Keqin Chen, Zhao Zhang, Weili Zeng, Richong Zhang, Feng Zhu, and Rui Zhao. 2023{\natexlab{a}}.
\newblock Shikra: Unleashing multimodal llm's referential dialogue magic.
\newblock \emph{arXiv preprint arXiv:2306.15195}.

\bibitem[{Chen et~al.(2024{\natexlab{c}})Chen, Li, Dong, Zhang, Zang, Chen, Duan, Wang, Qiao, Lin et~al.}]{chen2024we}
Lin Chen, Jinsong Li, Xiaoyi Dong, Pan Zhang, Yuhang Zang, Zehui Chen, Haodong Duan, Jiaqi Wang, Yu~Qiao, Dahua Lin, et~al. 2024{\natexlab{c}}.
\newblock Are we on the right way for evaluating large vision-language models?
\newblock \emph{arXiv preprint arXiv:2403.20330}.

\bibitem[{Chen et~al.(2023{\natexlab{b}})Chen, Li, Dong, Zhang, He, Wang, Zhao, and Lin}]{chen2023sharegpt4v}
Lin Chen, Jisong Li, Xiaoyi Dong, Pan Zhang, Conghui He, Jiaqi Wang, Feng Zhao, and Dahua Lin. 2023{\natexlab{b}}.
\newblock Sharegpt4v: Improving large multi-modal models with better captions.
\newblock \emph{arXiv preprint arXiv:2311.12793}.

\bibitem[{Chen et~al.(2024{\natexlab{d}})Chen, Wang, Tian, Ye, Gao, Cui, Tong, Hu, Luo, Ma et~al.}]{chen2024far}
Zhe Chen, Weiyun Wang, Hao Tian, Shenglong Ye, Zhangwei Gao, Erfei Cui, Wenwen Tong, Kongzhi Hu, Jiapeng Luo, Zheng Ma, et~al. 2024{\natexlab{d}}.
\newblock How far are we to gpt-4v? closing the gap to commercial multimodal models with open-source suites.
\newblock \emph{arXiv preprint arXiv:2404.16821}.

\bibitem[{Chen et~al.(2023{\natexlab{c}})Chen, Wu, Wang, Su, Chen, Xing, Muyan, Zhang, Zhu, Lu et~al.}]{chen2023internvl}
Zhe Chen, Jiannan Wu, Wenhai Wang, Weijie Su, Guo Chen, Sen Xing, Zhong Muyan, Qinglong Zhang, Xizhou Zhu, Lewei Lu, et~al. 2023{\natexlab{c}}.
\newblock Internvl: Scaling up vision foundation models and aligning for generic visual-linguistic tasks.
\newblock \emph{arXiv preprint arXiv:2312.14238}.

\bibitem[{Chu et~al.(2023)Chu, Qiao, Lin, Xu, Yang, Hu, Wei, Zhang, Zhang, Wei et~al.}]{chu2023mobilevlm}
Xiangxiang Chu, Limeng Qiao, Xinyang Lin, Shuang Xu, Yang Yang, Yiming Hu, Fei Wei, Xinyu Zhang, Bo~Zhang, Xiaolin Wei, et~al. 2023.
\newblock Mobilevlm: A fast, reproducible and strong vision language assistant for mobile devices.
\newblock \emph{arXiv preprint arXiv:2312.16886}.

\bibitem[{Chu et~al.(2024)Chu, Qiao, Zhang, Xu, Wei, Yang, Sun, Hu, Lin, Zhang et~al.}]{chu2024mobilevlm}
Xiangxiang Chu, Limeng Qiao, Xinyu Zhang, Shuang Xu, Fei Wei, Yang Yang, Xiaofei Sun, Yiming Hu, Xinyang Lin, Bo~Zhang, et~al. 2024.
\newblock Mobilevlm v2: Faster and stronger baseline for vision language model.
\newblock \emph{arXiv preprint arXiv:2402.03766}.

\bibitem[{Contributors(2023)}]{2023xtuner}
XTuner Contributors. 2023.
\newblock Xtuner: A toolkit for efficiently fine-tuning llm.
\newblock \url{https://github.com/InternLM/xtuner}.

\bibitem[{Cui et~al.(2023)Cui, Yang, and Yao}]{cui2023efficient}
Yiming Cui, Ziqing Yang, and Xin Yao. 2023.
\newblock Efficient and effective text encoding for chinese llama and alpaca.
\newblock \emph{arXiv preprint arXiv:2304.08177}.

\bibitem[{Dai et~al.(2023)Dai, Li, Li, Tiong, Zhao, Wang, Li, Fung, and Hoi}]{dai2023instructblip}
Wenliang Dai, Junnan Li, Dongxu Li, Anthony Tiong, Junqi Zhao, Weisheng Wang, Boyang Li, Pascale Fung, and Steven Hoi. 2023.
\newblock Instruct{BLIP}: Towards general-purpose vision-language models with instruction tuning.
\newblock In \emph{Thirty-seventh Conference on Neural Information Processing Systems}.

\bibitem[{Dao(2023)}]{dao2023flashattention}
Tri Dao. 2023.
\newblock Flashattention-2: Faster attention with better parallelism and work partitioning.
\newblock \emph{arXiv preprint arXiv:2307.08691}.

\bibitem[{Dao et~al.(2022)Dao, Fu, Ermon, Rudra, and R{\'e}}]{dao2022flashattention}
Tri Dao, Dan Fu, Stefano Ermon, Atri Rudra, and Christopher R{\'e}. 2022.
\newblock Flashattention: Fast and memory-efficient exact attention with io-awareness.
\newblock \emph{Advances in Neural Information Processing Systems}, 35:16344--16359.

\bibitem[{Dettmers et~al.(2023)Dettmers, Pagnoni, Holtzman, and Zettlemoyer}]{dettmers2023qlora}
Tim Dettmers, Artidoro Pagnoni, Ari Holtzman, and Luke Zettlemoyer. 2023.
\newblock Qlora: Efficient finetuning of quantized llms.
\newblock \emph{arXiv preprint arXiv:2305.14314}.

\bibitem[{Frantar and Alistarh(2023)}]{frantar2023sparsegpt}
Elias Frantar and Dan Alistarh. 2023.
\newblock \href {https://arxiv.org/abs/2301.00774} {Sparsegpt: Massive language models can be accurately pruned in one-shot}.
\newblock \emph{Preprint}, arXiv:2301.00774.

\bibitem[{Frantar et~al.(2022)Frantar, Ashkboos, Hoefler, and Alistarh}]{frantar2022gptq}
Elias Frantar, Saleh Ashkboos, Torsten Hoefler, and Dan Alistarh. 2022.
\newblock Gptq: Accurate post-training quantization for generative pre-trained transformers.
\newblock \emph{arXiv preprint arXiv:2210.17323}.

\bibitem[{Freitag and Al-Onaizan(2017)}]{freitag-al-onaizan-2017-beam}
Markus Freitag and Yaser Al-Onaizan. 2017.
\newblock \href {https://doi.org/10.18653/v1/W17-3207} {Beam search strategies for neural machine translation}.
\newblock In \emph{Proceedings of the First Workshop on Neural Machine Translation}, pages 56--60, Vancouver. Association for Computational Linguistics.

\bibitem[{Fu et~al.(2023)Fu, Chen, Shen, Qin, Zhang, Lin, Yang, Zheng, Li, Sun et~al.}]{fu2023mme}
Chaoyou Fu, Peixian Chen, Yunhang Shen, Yulei Qin, Mengdan Zhang, Xu~Lin, Jinrui Yang, Xiawu Zheng, Ke~Li, Xing Sun, et~al. 2023.
\newblock Mme: A comprehensive evaluation benchmark for multimodal large language models.
\newblock \emph{arXiv preprint arXiv:2306.13394}.

\bibitem[{Gao et~al.(2023)Gao, Pi, Zhang, Ye, Zhong, Wang, Hong, Han, Xu, Li et~al.}]{gao2023g}
Jiahui Gao, Renjie Pi, Jipeng Zhang, Jiacheng Ye, Wanjun Zhong, Yufei Wang, Lanqing Hong, Jianhua Han, Hang Xu, Zhenguo Li, et~al. 2023.
\newblock G-llava: Solving geometric problem with multi-modal large language model.
\newblock \emph{arXiv preprint arXiv:2312.11370}.

\bibitem[{Gao et~al.(2024)Gao, Zhang, Liu, Qiu, Huang, Lin, Zhao, Geng, Lin, Jin et~al.}]{gao2024sphinx}
Peng Gao, Renrui Zhang, Chris Liu, Longtian Qiu, Siyuan Huang, Weifeng Lin, Shitian Zhao, Shijie Geng, Ziyi Lin, Peng Jin, et~al. 2024.
\newblock Sphinx-x: Scaling data and parameters for a family of multi-modal large language models.
\newblock \emph{arXiv preprint arXiv:2402.05935}.

\bibitem[{Goncharova et~al.(2024)Goncharova, Razzhigaev, Mikhalchuk, Kurkin, Abdullaeva, Skripkin, Oseledets, Dimitrov, and Kuznetsov}]{goncharova2024omnifusion}
Elizaveta Goncharova, Anton Razzhigaev, Matvey Mikhalchuk, Maxim Kurkin, Irina Abdullaeva, Matvey Skripkin, Ivan Oseledets, Denis Dimitrov, and Andrey Kuznetsov. 2024.
\newblock Omnifusion technical report.
\newblock \emph{arXiv preprint arXiv:2404.06212}.

\bibitem[{Guo et~al.(2024)Guo, Cheng, Tang, and Lin}]{guo2024dynamic}
Yongxin Guo, Zhenglin Cheng, Xiaoying Tang, and Tao Lin. 2024.
\newblock Dynamic mixture of experts: An auto-tuning approach for efficient transformer models.
\newblock \emph{arXiv preprint arXiv:2405.14297}.

\bibitem[{He et~al.(2024)He, Liu, Wu, Yuan, Wang, Huang, and Zhao}]{he2024efficient}
Muyang He, Yexin Liu, Boya Wu, Jianhao Yuan, Yueze Wang, Tiejun Huang, and Bo~Zhao. 2024.
\newblock Efficient multimodal learning from data-centric perspective.
\newblock \emph{arXiv preprint arXiv:2402.11530}.

\bibitem[{Hendrycks and Gimpel(2016)}]{hendrycks2016gaussian}
Dan Hendrycks and Kevin Gimpel. 2016.
\newblock Gaussian error linear units (gelus).
\newblock \emph{arXiv preprint arXiv:1606.08415}.

\bibitem[{Hu et~al.(2024{\natexlab{a}})Hu, Xu, Ye, Yan, Zhang, Zhang, Li, Zhang, Jin, Huang et~al.}]{hu2024mplug}
Anwen Hu, Haiyang Xu, Jiabo Ye, Ming Yan, Liang Zhang, Bo~Zhang, Chen Li, Ji~Zhang, Qin Jin, Fei Huang, et~al. 2024{\natexlab{a}}.
\newblock mplug-docowl 1.5: Unified structure learning for ocr-free document understanding.
\newblock \emph{arXiv preprint arXiv:2403.12895}.

\bibitem[{Hu et~al.(2021)Hu, Shen, Wallis, Allen-Zhu, Li, Wang, Wang, and Chen}]{hu2021lora}
Edward~J Hu, Yelong Shen, Phillip Wallis, Zeyuan Allen-Zhu, Yuanzhi Li, Shean Wang, Lu~Wang, and Weizhu Chen. 2021.
\newblock Lora: Low-rank adaptation of large language models.
\newblock \emph{arXiv preprint arXiv:2106.09685}.

\bibitem[{Hu et~al.(2024{\natexlab{b}})Hu, Tu, Han, He, Cui, Long, Zheng, Fang, Huang, Zhao et~al.}]{hu2024minicpm}
Shengding Hu, Yuge Tu, Xu~Han, Chaoqun He, Ganqu Cui, Xiang Long, Zhi Zheng, Yewei Fang, Yuxiang Huang, Weilin Zhao, et~al. 2024{\natexlab{b}}.
\newblock Minicpm: Unveiling the potential of small language models with scalable training strategies.
\newblock \emph{arXiv preprint arXiv:2404.06395}.

\bibitem[{Jiao et~al.(2024)Jiao, Chen, Huang, Li, and Shen}]{jiao2024enhancing}
Qirui Jiao, Daoyuan Chen, Yilun Huang, Yaliang Li, and Ying Shen. 2024.
\newblock Enhancing multimodal large language models with vision detection models: An empirical study.
\newblock \emph{arXiv preprint arXiv:2401.17981}.

\bibitem[{Kalamkar et~al.(2019)Kalamkar, Mudigere, Mellempudi, Das, Banerjee, Avancha, Vooturi, Jammalamadaka, Huang, Yuen et~al.}]{kalamkar2019study}
Dhiraj Kalamkar, Dheevatsa Mudigere, Naveen Mellempudi, Dipankar Das, Kunal Banerjee, Sasikanth Avancha, Dharma~Teja Vooturi, Nataraj Jammalamadaka, Jianyu Huang, Hector Yuen, et~al. 2019.
\newblock A study of bfloat16 for deep learning training.
\newblock \emph{arXiv preprint arXiv:1905.12322}.

\bibitem[{Kar et~al.(2024)Kar, Tonioni, Poklukar, Kulshrestha, Zamir, and Tombari}]{kar2024brave}
O{\u{g}}uzhan~Fatih Kar, Alessio Tonioni, Petra Poklukar, Achin Kulshrestha, Amir Zamir, and Federico Tombari. 2024.
\newblock Brave: Broadening the visual encoding of vision-language models.
\newblock \emph{arXiv preprint arXiv:2404.07204}.

\bibitem[{Kembhavi et~al.(2016)Kembhavi, Salvato, Kolve, Seo, Hajishirzi, and Farhadi}]{kembhavi2016diagram}
Aniruddha Kembhavi, Mike Salvato, Eric Kolve, Minjoon Seo, Hannaneh Hajishirzi, and Ali Farhadi. 2016.
\newblock A diagram is worth a dozen images.
\newblock In \emph{Computer Vision--ECCV 2016: 14th European Conference, Amsterdam, The Netherlands, October 11--14, 2016, Proceedings, Part IV 14}, pages 235--251. Springer.

\bibitem[{Kim et~al.(2021)Kim, Lee, and Ro}]{kim2021distilling}
Junho Kim, Byung-Kwan Lee, and Yong~Man Ro. 2021.
\newblock Distilling robust and non-robust features in adversarial examples by information bottleneck.
\newblock \emph{Advances in Neural Information Processing Systems}, 34:17148--17159.

\bibitem[{Kim et~al.(2023{\natexlab{a}})Kim, Lee, and Ro}]{kim2023causal}
Junho Kim, Byung-Kwan Lee, and Yong~Man Ro. 2023{\natexlab{a}}.
\newblock Causal unsupervised semantic segmentation.
\newblock \emph{arXiv preprint arXiv:2310.07379}.

\bibitem[{Kim et~al.(2023{\natexlab{b}})Kim, Lee, and Ro}]{kim2023demystifying}
Junho Kim, Byung-Kwan Lee, and Yong~Man Ro. 2023{\natexlab{b}}.
\newblock Demystifying causal features on adversarial examples and causal inoculation for robust network by adversarial instrumental variable regression.
\newblock In \emph{Proceedings of the IEEE/CVF Conference on Computer Vision and Pattern Recognition}, pages 12302--12312.

\bibitem[{Kim et~al.(2023{\natexlab{c}})Kim, Hooper, Gholami, Dong, Li, Shen, Mahoney, and Keutzer}]{kim2023squeezellm}
Sehoon Kim, Coleman Hooper, Amir Gholami, Zhen Dong, Xiuyu Li, Sheng Shen, Michael~W Mahoney, and Kurt Keutzer. 2023{\natexlab{c}}.
\newblock Squeezellm: Dense-and-sparse quantization.
\newblock \emph{arXiv preprint arXiv:2306.07629}.

\bibitem[{Kim et~al.(2024)Kim, Kim, and Ro}]{kim2024improving}
Seongyeop Kim, Hyung-Il Kim, and Yong~Man Ro. 2024.
\newblock Improving open set recognition via visual prompts distilled from common-sense knowledge.
\newblock In \emph{Proceedings of the AAAI Conference on Artificial Intelligence}, volume~38, pages 2786--2794.

\bibitem[{Kim et~al.(2023{\natexlab{d}})Kim, Kim, Lee, Shin, and Ro}]{kim2023mitigating}
Yeonju Kim, Junho Kim, Byung-Kwan Lee, Sebin Shin, and Yong~Man Ro. 2023{\natexlab{d}}.
\newblock Mitigating dataset bias in image captioning through clip confounder-free captioning network.
\newblock In \emph{2023 IEEE International Conference on Image Processing (ICIP)}, pages 1720--1724. IEEE.

\bibitem[{Kwon et~al.(2023)Kwon, Li, Zhuang, Sheng, Zheng, Yu, Gonzalez, Zhang, and Stoica}]{kwon2023efficient}
Woosuk Kwon, Zhuohan Li, Siyuan Zhuang, Ying Sheng, Lianmin Zheng, Cody~Hao Yu, Joseph Gonzalez, Hao Zhang, and Ion Stoica. 2023.
\newblock Efficient memory management for large language model serving with pagedattention.
\newblock In \emph{Proceedings of the 29th Symposium on Operating Systems Principles}, pages 611--626.

\bibitem[{Lan et~al.(2020)Lan, Chen, Goodman, Gimpel, Sharma, and Soricut}]{lan2020albert}
Zhenzhong Lan, Mingda Chen, Sebastian Goodman, Kevin Gimpel, Piyush Sharma, and Radu Soricut. 2020.
\newblock \href {https://arxiv.org/abs/1909.11942} {Albert: A lite bert for self-supervised learning of language representations}.
\newblock \emph{Preprint}, arXiv:1909.11942.

\bibitem[{Lauren{\c{c}}on et~al.(2023)Lauren{\c{c}}on, Saulnier, Tronchon, Bekman, Singh, Lozhkov, Wang, Karamcheti, Rush, Kiela et~al.}]{laurenccon2023obelisc}
Hugo Lauren{\c{c}}on, Lucile Saulnier, L{\'e}o Tronchon, Stas Bekman, Amanpreet Singh, Anton Lozhkov, Thomas Wang, Siddharth Karamcheti, Alexander~M Rush, Douwe Kiela, et~al. 2023.
\newblock Obelisc: An open web-scale filtered dataset of interleaved image-text documents.
\newblock \emph{arXiv preprint arXiv:2306.16527}.

\bibitem[{Lee(2020)}]{lee2020training}
Byung-Kwan Lee. 2020.
\newblock Training encoder-attention through fully-connected crfs for efficient end-to-end lane detection model.

\bibitem[{Lee et~al.(2024{\natexlab{a}})Lee, Kim, Park, and Ro}]{lee2024meteor}
Byung-Kwan Lee, Chae~Won Kim, Beomchan Park, and Yong~Man Ro. 2024{\natexlab{a}}.
\newblock Meteor: Mamba-based traversal of rationale for large language and vision models.
\newblock \emph{arXiv preprint arXiv:2405.15574}.

\bibitem[{Lee et~al.(2022)Lee, Kim, and Ro}]{lee2022masking}
Byung-Kwan Lee, Junho Kim, and Yong~Man Ro. 2022.
\newblock Masking adversarial damage: Finding adversarial saliency for robust and sparse network.
\newblock In \emph{Proceedings of the IEEE/CVF Conference on Computer Vision and Pattern Recognition}, pages 15126--15136.

\bibitem[{Lee et~al.(2023)Lee, Kim, and Ro}]{lee2023mitigating}
Byung-Kwan Lee, Junho Kim, and Yong~Man Ro. 2023.
\newblock Mitigating adversarial vulnerability through causal parameter estimation by adversarial double machine learning.
\newblock In \emph{Proceedings of the IEEE/CVF International Conference on Computer Vision}, pages 4499--4509.

\bibitem[{Lee et~al.(2024{\natexlab{b}})Lee, Park, Kim, and Ro}]{lee2024collavo}
Byung-Kwan Lee, Beomchan Park, Chae~Won Kim, and Yong~Man Ro. 2024{\natexlab{b}}.
\newblock Collavo: Crayon large language and vision model.
\newblock \emph{arXiv preprint arXiv:2402.11248}.

\bibitem[{Lee et~al.(2024{\natexlab{c}})Lee, Park, Kim, and Ro}]{lee2024moai}
Byung-Kwan Lee, Beomchan Park, Chae~Won Kim, and Yong~Man Ro. 2024{\natexlab{c}}.
\newblock Moai: Mixture of all intelligence for large language and vision models.
\newblock \emph{arXiv preprint arXiv:2403.07508}.

\bibitem[{Lee et~al.(2021)Lee, Yu, and Ro}]{lee2020towards}
Byung-Kwan Lee, Youngjoon Yu, and Yong~Man Ro. 2021.
\newblock \href {https://openreview.net/forum?id=Cue2ZEBf12} {Towards adversarial robustness of bayesian neural network through hierarchical variational inference}.

\bibitem[{Li et~al.(2023{\natexlab{a}})Li, Zhang, Yang, Zhang, Pu, and Liu}]{li2023otterhd}
Bo~Li, Peiyuan Zhang, Jingkang Yang, Yuanhan Zhang, Fanyi Pu, and Ziwei Liu. 2023{\natexlab{a}}.
\newblock Otterhd: A high-resolution multi-modality model.
\newblock \emph{arXiv preprint arXiv:2311.04219}.

\bibitem[{Li et~al.(2023{\natexlab{b}})Li, Zhang, Chen, Wang, Yang, and Liu}]{li2023otter}
Bo~Li, Yuanhan Zhang, Liangyu Chen, Jinghao Wang, Jingkang Yang, and Ziwei Liu. 2023{\natexlab{b}}.
\newblock Otter: A multi-modal model with in-context instruction tuning.
\newblock \emph{arXiv preprint arXiv:2305.03726}.

\bibitem[{Li et~al.(2023{\natexlab{c}})Li, Wang, Wang, Ge, Ge, and Shan}]{li2023seed}
Bohao Li, Rui Wang, Guangzhi Wang, Yuying Ge, Yixiao Ge, and Ying Shan. 2023{\natexlab{c}}.
\newblock Seed-bench: Benchmarking multimodal llms with generative comprehension.
\newblock \emph{arXiv preprint arXiv:2307.16125}.

\bibitem[{Li et~al.(2024{\natexlab{a}})Li, Wang, Zhu, Kuo, Xu, Chen, Jain, Shi, and Wen}]{li2024cumo}
Jiachen Li, Xinyao Wang, Sijie Zhu, Chia-Wen Kuo, Lu~Xu, Fan Chen, Jitesh Jain, Humphrey Shi, and Longyin Wen. 2024{\natexlab{a}}.
\newblock Cumo: Scaling multimodal llm with co-upcycled mixture-of-experts.
\newblock \emph{arXiv preprint arXiv:2405.05949}.

\bibitem[{Li et~al.(2023{\natexlab{d}})Li, Li, Savarese, and Hoi}]{blip2}
Junnan Li, Dongxu Li, Silvio Savarese, and Steven Hoi. 2023{\natexlab{d}}.
\newblock Blip-2: Bootstrapping language-image pre-training with frozen image encoders and large language models.
\newblock \emph{arXiv preprint arXiv:2301.12597}.

\bibitem[{Li et~al.(2024{\natexlab{b}})Li, Zhang, Wang, Zhong, Chen, Chu, Liu, and Jia}]{li2024mini}
Yanwei Li, Yuechen Zhang, Chengyao Wang, Zhisheng Zhong, Yixin Chen, Ruihang Chu, Shaoteng Liu, and Jiaya Jia. 2024{\natexlab{b}}.
\newblock Mini-gemini: Mining the potential of multi-modality vision language models.
\newblock \emph{arXiv preprint arXiv:2403.18814}.

\bibitem[{Li et~al.(2023{\natexlab{e}})Li, Du, Zhou, Wang, Zhao, and Wen}]{li2023evaluating}
Yifan Li, Yifan Du, Kun Zhou, Jinpeng Wang, Wayne~Xin Zhao, and Ji-Rong Wen. 2023{\natexlab{e}}.
\newblock Evaluating object hallucination in large vision-language models.
\newblock \emph{arXiv preprint arXiv:2305.10355}.

\bibitem[{Li et~al.(2023{\natexlab{f}})Li, Yu, Liang, He, Karampatziakis, Chen, and Zhao}]{li2023loftq}
Yixiao Li, Yifan Yu, Chen Liang, Pengcheng He, Nikos Karampatziakis, Weizhu Chen, and Tuo Zhao. 2023{\natexlab{f}}.
\newblock \href {https://arxiv.org/abs/2310.08659} {Loftq: Lora-fine-tuning-aware quantization for large language models}.
\newblock \emph{Preprint}, arXiv:2310.08659.

\bibitem[{Li et~al.(2024{\natexlab{c}})Li, Jiang, Hu, Wang, Zhong, Luo, Ma, and Zhang}]{li2024uni}
Yunxin Li, Shenyuan Jiang, Baotian Hu, Longyue Wang, Wanqi Zhong, Wenhan Luo, Lin Ma, and Min Zhang. 2024{\natexlab{c}}.
\newblock Uni-moe: Scaling unified multimodal llms with mixture of experts.
\newblock \emph{arXiv preprint arXiv:2405.11273}.

\bibitem[{Li et~al.(2023{\natexlab{g}})Li, Yang, Liu, Ma, Zhang, Yang, Sun, Liu, and Bai}]{li2023monkey}
Zhang Li, Biao Yang, Qiang Liu, Zhiyin Ma, Shuo Zhang, Jingxu Yang, Yabo Sun, Yuliang Liu, and Xiang Bai. 2023{\natexlab{g}}.
\newblock Monkey: Image resolution and text label are important things for large multi-modal models.
\newblock \emph{arXiv preprint arXiv:2311.06607}.

\bibitem[{Lin et~al.(2024)Lin, Tang, Ye, Cui, Zhu, Jin, Zhang, Ning, and Yuan}]{lin2024moe}
Bin Lin, Zhenyu Tang, Yang Ye, Jiaxi Cui, Bin Zhu, Peng Jin, Junwu Zhang, Munan Ning, and Li~Yuan. 2024.
\newblock Moe-llava: Mixture of experts for large vision-language models.
\newblock \emph{arXiv preprint arXiv:2401.15947}.

\bibitem[{Lin et~al.(2023{\natexlab{a}})Lin, Tang, Tang, Yang, Chen, Wang, Xiao, Dang, Gan, and Han}]{lin2023awq}
Ji~Lin, Jiaming Tang, Haotian Tang, Shang Yang, Wei-Ming Chen, Wei-Chen Wang, Guangxuan Xiao, Xingyu Dang, Chuang Gan, and Song Han. 2023{\natexlab{a}}.
\newblock Awq: Activation-aware weight quantization for llm compression and acceleration.
\newblock \emph{arXiv preprint arXiv:2306.00978}.

\bibitem[{Lin et~al.(2023{\natexlab{b}})Lin, Yin, Ping, Lu, Molchanov, Tao, Mao, Kautz, Shoeybi, and Han}]{lin2023vila}
Ji~Lin, Hongxu Yin, Wei Ping, Yao Lu, Pavlo Molchanov, Andrew Tao, Huizi Mao, Jan Kautz, Mohammad Shoeybi, and Song Han. 2023{\natexlab{b}}.
\newblock Vila: On pre-training for visual language models.
\newblock \emph{arXiv preprint arXiv:2312.07533}.

\bibitem[{Lin et~al.(2023{\natexlab{c}})Lin, Liu, Zhang, Gao, Qiu, Xiao, Qiu, Lin, Shao, Chen et~al.}]{lin2023sphinx}
Ziyi Lin, Chris Liu, Renrui Zhang, Peng Gao, Longtian Qiu, Han Xiao, Han Qiu, Chen Lin, Wenqi Shao, Keqin Chen, et~al. 2023{\natexlab{c}}.
\newblock Sphinx: The joint mixing of weights, tasks, and visual embeddings for multi-modal large language models.
\newblock \emph{arXiv preprint arXiv:2311.07575}.

\bibitem[{Liu et~al.(2023{\natexlab{a}})Liu, Guan, Li, Chen, Yacoob, Manocha, and Zhou}]{liu2023hallusionbench}
Fuxiao Liu, Tianrui Guan, Zongxia Li, Lichang Chen, Yaser Yacoob, Dinesh Manocha, and Tianyi Zhou. 2023{\natexlab{a}}.
\newblock Hallusionbench: You see what you think? or you think what you see? an image-context reasoning benchmark challenging for gpt-4v (ision), llava-1.5, and other multi-modality models.
\newblock \emph{arXiv preprint arXiv:2310.14566}.

\bibitem[{Liu et~al.(2023{\natexlab{b}})Liu, Li, Li, and Lee}]{liu2023improved}
Haotian Liu, Chunyuan Li, Yuheng Li, and Yong~Jae Lee. 2023{\natexlab{b}}.
\newblock Improved baselines with visual instruction tuning.
\newblock \emph{arXiv preprint arXiv:2310.03744}.

\bibitem[{Liu et~al.(2024{\natexlab{a}})Liu, Li, Li, Li, Zhang, Shen, and Lee}]{liu2024llavanext}
Haotian Liu, Chunyuan Li, Yuheng Li, Bo~Li, Yuanhan Zhang, Sheng Shen, and Yong~Jae Lee. 2024{\natexlab{a}}.
\newblock \href {https://llava-vl.github.io/blog/2024-01-30-llava-next/} {Llava-next: Improved reasoning, ocr, and world knowledge}.

\bibitem[{Liu et~al.(2023{\natexlab{c}})Liu, Li, Wu, and Lee}]{liu2023visual}
Haotian Liu, Chunyuan Li, Qingyang Wu, and Yong~Jae Lee. 2023{\natexlab{c}}.
\newblock Visual instruction tuning.
\newblock In \emph{Thirty-seventh Conference on Neural Information Processing Systems}.

\bibitem[{Liu et~al.(2024{\natexlab{b}})Liu, Song, Lin, Lam, Neubig, Li, and Yue}]{liu2024visualwebbench}
Junpeng Liu, Yifan Song, Bill~Yuchen Lin, Wai Lam, Graham Neubig, Yuanzhi Li, and Xiang Yue. 2024{\natexlab{b}}.
\newblock Visualwebbench: How far have multimodal llms evolved in web page understanding and grounding?
\newblock \emph{arXiv preprint arXiv:2404.05955}.

\bibitem[{Liu et~al.(2023{\natexlab{d}})Liu, Duan, Zhang, Li, Zhang, Zhao, Yuan, Wang, He, Liu et~al.}]{liu2023mmbench}
Yuan Liu, Haodong Duan, Yuanhan Zhang, Bo~Li, Songyang Zhang, Wangbo Zhao, Yike Yuan, Jiaqi Wang, Conghui He, Ziwei Liu, et~al. 2023{\natexlab{d}}.
\newblock Mmbench: Is your multi-modal model an all-around player?
\newblock \emph{arXiv preprint arXiv:2307.06281}.

\bibitem[{Loshchilov and Hutter(2019)}]{loshchilov2018decoupled}
Ilya Loshchilov and Frank Hutter. 2019.
\newblock \href {https://openreview.net/forum?id=Bkg6RiCqY7} {Decoupled weight decay regularization}.
\newblock In \emph{International Conference on Learning Representations}.

\bibitem[{Lu et~al.(2024)Lu, Liu, Zhang, Wang, Dong, Liu, Sun, Ren, Li, Sun et~al.}]{lu2024deepseek}
Haoyu Lu, Wen Liu, Bo~Zhang, Bingxuan Wang, Kai Dong, Bo~Liu, Jingxiang Sun, Tongzheng Ren, Zhuoshu Li, Yaofeng Sun, et~al. 2024.
\newblock Deepseek-vl: towards real-world vision-language understanding.
\newblock \emph{arXiv preprint arXiv:2403.05525}.

\bibitem[{Lu et~al.(2023{\natexlab{a}})Lu, Clark, Lee, Zhang, Khosla, Marten, Hoiem, and Kembhavi}]{lu2023unifiedio}
Jiasen Lu, Christopher Clark, Sangho Lee, Zichen Zhang, Savya Khosla, Ryan Marten, Derek Hoiem, and Aniruddha Kembhavi. 2023{\natexlab{a}}.
\newblock \href {https://arxiv.org/abs/2312.17172} {Unified-io 2: Scaling autoregressive multimodal models with vision, language, audio, and action}.
\newblock \emph{Preprint}, arXiv:2312.17172.

\bibitem[{Lu et~al.(2023{\natexlab{b}})Lu, Bansal, Xia, Liu, Li, Hajishirzi, Cheng, Chang, Galley, and Gao}]{lu2023mathvista}
Pan Lu, Hritik Bansal, Tony Xia, Jiacheng Liu, Chunyuan Li, Hannaneh Hajishirzi, Hao Cheng, Kai-Wei Chang, Michel Galley, and Jianfeng Gao. 2023{\natexlab{b}}.
\newblock Mathvista: Evaluating mathematical reasoning of foundation models in visual contexts.
\newblock \emph{arXiv preprint arXiv:2310.02255}.

\bibitem[{Lu et~al.(2022)Lu, Mishra, Xia, Qiu, Chang, Zhu, Tafjord, Clark, and Kalyan}]{lu2022learn}
Pan Lu, Swaroop Mishra, Tanglin Xia, Liang Qiu, Kai-Wei Chang, Song-Chun Zhu, Oyvind Tafjord, Peter Clark, and Ashwin Kalyan. 2022.
\newblock Learn to explain: Multimodal reasoning via thought chains for science question answering.
\newblock \emph{Advances in Neural Information Processing Systems}, 35:2507--2521.

\bibitem[{Ma et~al.(2023)Ma, Fang, and Wang}]{ma2023llmpruner}
Xinyin Ma, Gongfan Fang, and Xinchao Wang. 2023.
\newblock \href {https://arxiv.org/abs/2305.11627} {Llm-pruner: On the structural pruning of large language models}.
\newblock \emph{Preprint}, arXiv:2305.11627.

\bibitem[{Masry et~al.(2022)Masry, Long, Tan, Joty, and Hoque}]{masry2022chartqa}
Ahmed Masry, Do~Xuan Long, Jia~Qing Tan, Shafiq Joty, and Enamul Hoque. 2022.
\newblock Chartqa: A benchmark for question answering about charts with visual and logical reasoning.
\newblock \emph{arXiv preprint arXiv:2203.10244}.

\bibitem[{McKinzie et~al.(2024)McKinzie, Gan, Fauconnier, Dodge, Zhang, Dufter, Shah, Du, Peng, Weers et~al.}]{mckinzie2024mm1}
Brandon McKinzie, Zhe Gan, Jean-Philippe Fauconnier, Sam Dodge, Bowen Zhang, Philipp Dufter, Dhruti Shah, Xianzhi Du, Futang Peng, Floris Weers, et~al. 2024.
\newblock Mm1: Methods, analysis \& insights from multimodal llm pre-training.
\newblock \emph{arXiv preprint arXiv:2403.09611}.

\bibitem[{Men et~al.(2024)Men, Xu, Zhang, Wang, Lin, Lu, Han, and Chen}]{men2024shortgpt}
Xin Men, Mingyu Xu, Qingyu Zhang, Bingning Wang, Hongyu Lin, Yaojie Lu, Xianpei Han, and Weipeng Chen. 2024.
\newblock \href {https://arxiv.org/abs/2403.03853} {Shortgpt: Layers in large language models are more redundant than you expect}.
\newblock \emph{Preprint}, arXiv:2403.03853.

\bibitem[{Park et~al.(2024{\natexlab{a}})Park, Park, Kim, Lee, Kim, Kwon, Kwon, Kim, Lee, and Lee}]{park2024lutgemm}
Gunho Park, Baeseong Park, Minsub Kim, Sungjae Lee, Jeonghoon Kim, Beomseok Kwon, Se~Jung Kwon, Byeongwook Kim, Youngjoo Lee, and Dongsoo Lee. 2024{\natexlab{a}}.
\newblock \href {https://arxiv.org/abs/2206.09557} {Lut-gemm: Quantized matrix multiplication based on luts for efficient inference in large-scale generative language models}.
\newblock \emph{Preprint}, arXiv:2206.09557.

\bibitem[{Park et~al.(2024{\natexlab{b}})Park, Kim, and Ro}]{park2024integrating}
Sungjune Park, Hyunjun Kim, and Yong~Man Ro. 2024{\natexlab{b}}.
\newblock Integrating language-derived appearance elements with visual cues in pedestrian detection.
\newblock \emph{IEEE Transactions on Circuits and Systems for Video Technology}.

\bibitem[{Park et~al.(2024{\natexlab{c}})Park, Kim, and Ro}]{park2024robust}
Sungjune Park, Hyunjun Kim, and Yong~Man Ro. 2024{\natexlab{c}}.
\newblock Robust pedestrian detection via constructing versatile pedestrian knowledge bank.
\newblock \emph{Pattern Recognition}, 153:110539.

\bibitem[{Radford et~al.(2021)Radford, Kim, Hallacy, Ramesh, Goh, Agarwal, Sastry, Askell, Mishkin, Clark, Krueger, and Sutskever}]{clip}
Alec Radford, Jong~Wook Kim, Chris Hallacy, Aditya Ramesh, Gabriel Goh, Sandhini Agarwal, Girish Sastry, Amanda Askell, Pamela Mishkin, Jack Clark, Gretchen Krueger, and Ilya Sutskever. 2021.
\newblock Learning transferable visual models from natural language supervision.
\newblock In \emph{Proceedings of the 38th International Conference on Machine Learning}, volume 139 of \emph{Proceedings of Machine Learning Research}, pages 8748--8763. PMLR.

\bibitem[{Ranzinger et~al.(2023)Ranzinger, Heinrich, Kautz, and Molchanov}]{ranzinger2023radio}
Mike Ranzinger, Greg Heinrich, Jan Kautz, and Pavlo Molchanov. 2023.
\newblock Am-radio: Agglomerative model--reduce all domains into one.
\newblock \emph{arXiv preprint arXiv:2312.06709}.

\bibitem[{Reid et~al.(2021)Reid, Marrese-Taylor, and Matsuo}]{reid2021subformer}
Machel Reid, Edison Marrese-Taylor, and Yutaka Matsuo. 2021.
\newblock \href {https://arxiv.org/abs/2101.00234} {Subformer: Exploring weight sharing for parameter efficiency in generative transformers}.
\newblock \emph{Preprint}, arXiv:2101.00234.

\bibitem[{Shao et~al.(2024{\natexlab{a}})Shao, Chen, Zhang, Xu, Zhao, Li, Zhang, Gao, Qiao, and Luo}]{shao2024omniquant}
Wenqi Shao, Mengzhao Chen, Zhaoyang Zhang, Peng Xu, Lirui Zhao, Zhiqian Li, Kaipeng Zhang, Peng Gao, Yu~Qiao, and Ping Luo. 2024{\natexlab{a}}.
\newblock \href {https://arxiv.org/abs/2308.13137} {Omniquant: Omnidirectionally calibrated quantization for large language models}.
\newblock \emph{Preprint}, arXiv:2308.13137.

\bibitem[{Shao et~al.(2024{\natexlab{b}})Shao, Yu, Yu, Ouyang, Zheng, Gai, Wang, and Ding}]{shao2024imp}
Zhenwei Shao, Zhou Yu, Jun Yu, Xuecheng Ouyang, Lihao Zheng, Zhenbiao Gai, Mingyang Wang, and Jiajun Ding. 2024{\natexlab{b}}.
\newblock Imp: Highly capable large multimodal models for mobile devices.
\newblock \emph{arXiv preprint arXiv:2405.12107}.

\bibitem[{Shazeer et~al.(2017)Shazeer, Mirhoseini, Maziarz, Davis, Le, Hinton, and Dean}]{shazeer2017}
Noam Shazeer, *Azalia Mirhoseini, *Krzysztof Maziarz, Andy Davis, Quoc Le, Geoffrey Hinton, and Jeff Dean. 2017.
\newblock \href {https://openreview.net/forum?id=B1ckMDqlg} {Outrageously large neural networks: The sparsely-gated mixture-of-experts layer}.
\newblock In \emph{International Conference on Learning Representations}.

\bibitem[{Sohoni et~al.(2019)Sohoni, Aberger, Leszczynski, Zhang, and R{\'e}}]{sohoni2019low}
Nimit~S Sohoni, Christopher~R Aberger, Megan Leszczynski, Jian Zhang, and Christopher R{\'e}. 2019.
\newblock Low-memory neural network training: A technical report.
\newblock \emph{arXiv preprint arXiv:1904.10631}.

\bibitem[{Sun et~al.(2024{\natexlab{a}})Sun, Zhou, Li, Lu, Yi, Chen, Xu, Luo, Zhang, Zhan et~al.}]{sun2024parrot}
Hai-Long Sun, Da-Wei Zhou, Yang Li, Shiyin Lu, Chao Yi, Qing-Guo Chen, Zhao Xu, Weihua Luo, Kaifu Zhang, De-Chuan Zhan, et~al. 2024{\natexlab{a}}.
\newblock Parrot: Multilingual visual instruction tuning.
\newblock \emph{arXiv preprint arXiv:2406.02539}.

\bibitem[{Sun et~al.(2024{\natexlab{b}})Sun, Liu, Bair, and Kolter}]{sun2024simple}
Mingjie Sun, Zhuang Liu, Anna Bair, and J.~Zico Kolter. 2024{\natexlab{b}}.
\newblock \href {https://arxiv.org/abs/2306.11695} {A simple and effective pruning approach for large language models}.
\newblock \emph{Preprint}, arXiv:2306.11695.

\bibitem[{Sun et~al.(2023)Sun, Cui, Zhang, Zhang, Yu, Luo, Wang, Rao, Liu, Huang et~al.}]{sun2023generative}
Quan Sun, Yufeng Cui, Xiaosong Zhang, Fan Zhang, Qiying Yu, Zhengxiong Luo, Yueze Wang, Yongming Rao, Jingjing Liu, Tiejun Huang, et~al. 2023.
\newblock Generative multimodal models are in-context learners.
\newblock \emph{arXiv preprint arXiv:2312.13286}.

\bibitem[{Takase and Kiyono(2023)}]{takase2023lessons}
Sho Takase and Shun Kiyono. 2023.
\newblock \href {https://arxiv.org/abs/2104.06022} {Lessons on parameter sharing across layers in transformers}.
\newblock \emph{Preprint}, arXiv:2104.06022.

\bibitem[{Team et~al.(2023)Team, Anil, Borgeaud, Wu, Alayrac, Yu, Soricut, Schalkwyk, Dai, Hauth et~al.}]{team2023gemini}
Gemini Team, Rohan Anil, Sebastian Borgeaud, Yonghui Wu, Jean-Baptiste Alayrac, Jiahui Yu, Radu Soricut, Johan Schalkwyk, Andrew~M Dai, Anja Hauth, et~al. 2023.
\newblock Gemini: a family of highly capable multimodal models.
\newblock \emph{arXiv preprint arXiv:2312.11805}.

\bibitem[{Team(2023)}]{2023internlm}
InternLM Team. 2023.
\newblock Internlm: A multilingual language model with progressively enhanced capabilities.
\newblock \url{https://github.com/InternLM/InternLM-techreport}.

\bibitem[{Thawakar et~al.(2024)Thawakar, Vayani, Khan, Cholakal, Anwer, Felsberg, Baldwin, Xing, and Khan}]{thawakar2024mobillama}
Omkar Thawakar, Ashmal Vayani, Salman Khan, Hisham Cholakal, Rao~M. Anwer, Michael Felsberg, Tim Baldwin, Eric~P. Xing, and Fahad~Shahbaz Khan. 2024.
\newblock \href {https://arxiv.org/abs/2402.16840} {Mobillama: Towards accurate and lightweight fully transparent gpt}.
\newblock \emph{Preprint}, arXiv:2402.16840.

\bibitem[{Wang et~al.(2024{\natexlab{a}})Wang, Pan, Shi, Lu, Zhan, and Li}]{wang2024measuring}
Ke~Wang, Junting Pan, Weikang Shi, Zimu Lu, Mingjie Zhan, and Hongsheng Li. 2024{\natexlab{a}}.
\newblock Measuring multimodal mathematical reasoning with math-vision dataset.
\newblock \emph{arXiv preprint arXiv:2402.14804}.

\bibitem[{Wang et~al.(2023)Wang, Lv, Yu, Hong, Qi, Wang, Ji, Yang, Zhao, Song et~al.}]{wang2023cogvlm}
Weihan Wang, Qingsong Lv, Wenmeng Yu, Wenyi Hong, Ji~Qi, Yan Wang, Junhui Ji, Zhuoyi Yang, Lei Zhao, Xixuan Song, et~al. 2023.
\newblock Cogvlm: Visual expert for pretrained language models.
\newblock \emph{arXiv preprint arXiv:2311.03079}.

\bibitem[{Wang et~al.(2024{\natexlab{b}})Wang, Ren, Luo, Li, Yan, Chen, Wang, Li, Lu, Zhu et~al.}]{wang2024all}
Weiyun Wang, Yiming Ren, Haowen Luo, Tiantong Li, Chenxiang Yan, Zhe Chen, Wenhai Wang, Qingyun Li, Lewei Lu, Xizhou Zhu, et~al. 2024{\natexlab{b}}.
\newblock The all-seeing project v2: Towards general relation comprehension of the open world.
\newblock \emph{arXiv preprint arXiv:2402.19474}.

\bibitem[{Wu et~al.(2023)Wu, Zhang, Zhang, Chen, Liao, Wang, Li, Sun, Yan, Zhai et~al.}]{wu2023q}
Haoning Wu, Zicheng Zhang, Erli Zhang, Chaofeng Chen, Liang Liao, Annan Wang, Chunyi Li, Wenxiu Sun, Qiong Yan, Guangtao Zhai, et~al. 2023.
\newblock Q-bench: A benchmark for general-purpose foundation models on low-level vision.
\newblock \emph{arXiv preprint arXiv:2309.14181}.

\bibitem[{Xu et~al.(2024)Xu, Yao, Guo, Cui, Ni, Ge, Chua, Liu, Sun, and Huang}]{xu2024llava}
Ruyi Xu, Yuan Yao, Zonghao Guo, Junbo Cui, Zanlin Ni, Chunjiang Ge, Tat-Seng Chua, Zhiyuan Liu, Maosong Sun, and Gao Huang. 2024.
\newblock Llava-uhd: an lmm perceiving any aspect ratio and high-resolution images.
\newblock \emph{arXiv preprint arXiv:2403.11703}.

\bibitem[{Xue et~al.(2024)Xue, Song, Mi, Chen, Xia, and Chen}]{xue2024powerinfer2}
Zhenliang Xue, Yixin Song, Zeyu Mi, Le~Chen, Yubin Xia, and Haibo Chen. 2024.
\newblock \href {https://arxiv.org/abs/2406.06282} {Powerinfer-2: Fast large language model inference on a smartphone}.

\bibitem[{Ye et~al.(2024)Ye, Tao, Huang, and Li}]{ye2024chunkattention}
Lu~Ye, Ze~Tao, Yong Huang, and Yang Li. 2024.
\newblock Chunkattention: Efficient self-attention with prefix-aware kv cache and two-phase partition.
\newblock \emph{arXiv preprint arXiv:2402.15220}.

\bibitem[{Ye et~al.(2023{\natexlab{a}})Ye, Xu, Xu, Ye, Yan, Zhou, Wang, Hu, Shi, Shi et~al.}]{ye2023mplug}
Qinghao Ye, Haiyang Xu, Guohai Xu, Jiabo Ye, Ming Yan, Yiyang Zhou, Junyang Wang, Anwen Hu, Pengcheng Shi, Yaya Shi, et~al. 2023{\natexlab{a}}.
\newblock mplug-owl: Modularization empowers large language models with multimodality.
\newblock \emph{arXiv preprint arXiv:2304.14178}.

\bibitem[{Ye et~al.(2023{\natexlab{b}})Ye, Xu, Ye, Yan, Liu, Qian, Zhang, Huang, and Zhou}]{ye2023mplug2}
Qinghao Ye, Haiyang Xu, Jiabo Ye, Ming Yan, Haowei Liu, Qi~Qian, Ji~Zhang, Fei Huang, and Jingren Zhou. 2023{\natexlab{b}}.
\newblock mplug-owl2: Revolutionizing multi-modal large language model with modality collaboration.
\newblock \emph{arXiv preprint arXiv:2311.04257}.

\bibitem[{Young et~al.(2024)Young, Chen, Li, Huang, Zhang, Zhang, Li, Zhu, Chen, Chang et~al.}]{young2024yi}
Alex Young, Bei Chen, Chao Li, Chengen Huang, Ge~Zhang, Guanwei Zhang, Heng Li, Jiangcheng Zhu, Jianqun Chen, Jing Chang, et~al. 2024.
\newblock Yi: Open foundation models by 01. ai.
\newblock \emph{arXiv preprint arXiv:2403.04652}.

\bibitem[{Yu et~al.(2023)Yu, Yang, Li, Wang, Lin, Liu, Wang, and Wang}]{yu2023mm}
Weihao Yu, Zhengyuan Yang, Linjie Li, Jianfeng Wang, Kevin Lin, Zicheng Liu, Xinchao Wang, and Lijuan Wang. 2023.
\newblock Mm-vet: Evaluating large multimodal models for integrated capabilities.
\newblock \emph{arXiv preprint arXiv:2308.02490}.

\bibitem[{Yue et~al.(2023)Yue, Qu, Zhang, Fu, Huang, Sun, Su, and Chen}]{yue2023mammoth}
Xiang Yue, Xingwei Qu, Ge~Zhang, Yao Fu, Wenhao Huang, Huan Sun, Yu~Su, and Wenhu Chen. 2023.
\newblock Mammoth: Building math generalist models through hybrid instruction tuning.
\newblock \emph{arXiv preprint arXiv:2309.05653}.

\bibitem[{Yue et~al.(2024)Yue, Zheng, Zhang, and Chen}]{yue2024mammoth2}
Xiang Yue, Tuney Zheng, Ge~Zhang, and Wenhu Chen. 2024.
\newblock \href {https://arxiv.org/abs/2405.03548} {Mammoth2: Scaling instructions from the web}.

\bibitem[{Zhang et~al.(2023)Zhang, Wang, Cao, Xu, Ouyang, Zhao, Ding, Zhang, Duan, Yan et~al.}]{zhang2023internlm}
Pan Zhang, Xiaoyi Dong~Bin Wang, Yuhang Cao, Chao Xu, Linke Ouyang, Zhiyuan Zhao, Shuangrui Ding, Songyang Zhang, Haodong Duan, Hang Yan, et~al. 2023.
\newblock Internlm-xcomposer: A vision-language large model for advanced text-image comprehension and composition.
\newblock \emph{arXiv preprint arXiv:2309.15112}.

\bibitem[{Zhang et~al.(2024)Zhang, Jiang, Zhang, Lin, Guo, Qiu, Zhou, Lu, Chang, Gao et~al.}]{zhang2024mathverse}
Renrui Zhang, Dongzhi Jiang, Yichi Zhang, Haokun Lin, Ziyu Guo, Pengshuo Qiu, Aojun Zhou, Pan Lu, Kai-Wei Chang, Peng Gao, et~al. 2024.
\newblock Mathverse: Does your multi-modal llm truly see the diagrams in visual math problems?
\newblock \emph{arXiv preprint arXiv:2403.14624}.

\bibitem[{Zhao et~al.(2024)Zhao, Zhang, Zhao, Ding, Huang, and Wang}]{zhao2024cobra}
Han Zhao, Min Zhang, Wei Zhao, Pengxiang Ding, Siteng Huang, and Donglin Wang. 2024.
\newblock Cobra: Extending mamba to multi-modal large language model for efficient inference.
\newblock \emph{arXiv preprint arXiv:2403.14520}.

\bibitem[{Zhou et~al.(2024)Zhou, Hu, Weng, Jia, Luo, Liu, Wu, and Huang}]{zhou2024tinyllava}
Baichuan Zhou, Ying Hu, Xi~Weng, Junlong Jia, Jie Luo, Xien Liu, Ji~Wu, and Lei Huang. 2024.
\newblock Tinyllava: A framework of small-scale large multimodal models.
\newblock \emph{arXiv preprint arXiv:2402.14289}.

\bibitem[{Zhu et~al.(2023)Zhu, Chen, Shen, Li, and Elhoseiny}]{zhu2023minigpt}
Deyao Zhu, Jun Chen, Xiaoqian Shen, Xiang Li, and Mohamed Elhoseiny. 2023.
\newblock Minigpt-4: Enhancing vision-language understanding with advanced large language models.
\newblock \emph{arXiv preprint arXiv:2304.10592}.

\bibitem[{Zhu et~al.(2024)Zhu, Zhu, Liu, Ou, Mou, and Tang}]{zhu2024llava}
Yichen Zhu, Minjie Zhu, Ning Liu, Zhicai Ou, Xiaofeng Mou, and Jian Tang. 2024.
\newblock Llava-phi: Efficient multi-modal assistant with small language model.
\newblock \emph{arXiv preprint arXiv:2401.02330}.

\end{thebibliography}

\clearpage
\appendix
\onecolumn

\section{Evaluation Benchmarks}
\label{app:A}

\begin{itemize}
    \item \textbf{Q-Bench}~\citep{wu2023q} is designed to evaluate the low-level visual abilities of Multi-modality Large Language Models (MLLMs). It is segmented into three primary categories: perception, description, and assessment. The perception section focuses on the ability of MLLMs to identify and interpret basic image attributes. The description section checks the precision and completeness of how MLLMs can articulate these attributes. The assessment section measures the extent to which MLLMs' evaluations of image quality match human judgments. The dataset contains a total of 81,284 samples.
    
    \item \textbf{SQA-IMG (SQA$^{\text{I}}$)}~\citep{lu2022learn} is part of the broader ScienceQA (SQA) dataset, which aims to improve reasoning and interpretability in AI systems through science-based question answering. This dataset covers a wide range of science disciplines, featuring 26 different topics in natural, social, and language sciences, all accompanied by annotated answers, lectures, and explanations. SQA-IMG includes image-related samples, amounting to 10,332 question-answer pairs.
    
    \item \textbf{AI2D}~\citep{kembhavi2016diagram} or AI2 Diagrams, addresses diagram interpretation and reasoning challenges, focusing on syntactic parsing and semantic understanding. It supports research into diagram structure and element relationships, critical for tasks like diagram-based question answering. This collection includes over 5,000 diagrams from elementary science topics, along with over 15,000 multiple-choice questions.
    
    \item \textbf{ChartQA}~\citep{masry2022chartqa} develops to challenge and improve question answering systems that deal with data visualizations like bar charts, line charts, and pie charts. This dataset tests systems on questions requiring arithmetic and logical reasoning and includes both human-generated and machine-created question-answer pairs. It comprises 32,719 samples in total.
    
    \item \textbf{SEED-IMG (SEED$^{\text{I}}$)}~\citep{li2023seed}, a subset of SEED-Bench, evaluates the generative comprehension skills of multimodal large language models (MLLMs) with a focus on spatial and temporal understanding. It offers several subsets mapped to 12 evaluation dimensions across image and video modalities, with SEED-IMG specifically concentrating on images.
    
    \item \textbf{POPE}~\citep{li2023evaluating} introduces a method to systematically assess the tendency of LLVMs to falsely generate nonexistent objects in images. This method turns the evaluation into a binary classification task using polling questions, providing a fair and adaptable approach.
    
    \item \textbf{HallusionBench (HallB)}~\citep{liu2023hallusionbench} is crafted to evaluate and explore visual illusions and knowledge hallucinations in large language and vision models (LLVMs). This benchmark uses carefully crafted example pairs to identify model failures, featuring diverse visual-question pairs including subsets focused on illusions, math, charts, tables, maps, and OCR. It includes 346 images and 1,129 questions.
    
    \item \textbf{MME}~\citep{fu2023mme} serves as a comprehensive evaluation framework for Multimodal Large Language Models (MLLMs), focusing on various perception and cognition tasks through 14 sub-tasks like coarse and fine-grained recognition, OCR, and commonsense reasoning. This benchmark aims to address existing evaluation gaps and ensures a thorough testing environment for MLLMs.
    
    \item \textbf{MathVista}~\citep{lu2023mathvista} is an extensive benchmark designed to test visual-based mathematical reasoning in AI models. It integrates visual understanding in evaluating models' abilities to solve math problems that involve visuals. The dataset consists of three subsets: IQTest, FunctionQA, and PaperQA, totaling 6,141 examples.
    
    \item \textbf{MMB, MMB-Chinese (MMB$^{\text{CN}}$)}~\citep{liu2023mmbench} aims to establish a robust evaluation standard for vision language models by covering a broad spectrum of necessary multimodal comprehension skills (20 fine-grained abilities) in both English and Chinese. This benchmark consists of 3,217 questions gathered from various sources to challenge different facets of LLVMs.
    
    \item \textbf{MM-Vet}~\citep{yu2023mm} is designed to systematically evaluate LMMs on complex tasks requiring multiple vision language (VL) capabilities. It tests recognition, knowledge, OCR, spatial awareness, language generation, and math, integrating these abilities into 16 different task combinations. The dataset includes 200 images and 218 questions, each requiring the integration of multiple capabilities.
    
    \item \textbf{LLaVA Bench in the Wild (LLaVA$^\text{W}$)}~\citep{liu2023visual} assesses large multimodal models (LMM) on complex tasks and new domains through a collection of 24 images with 60 questions. This dataset features diverse settings, including indoor, outdoor, artworks, and memes, with each image accompanied by detailed descriptions and curated questions.
    
    \item \textbf{MMStar}~\citep{chen2024we} is crafted to precisely evaluate the true multimodal capabilities of LLVMs by ensuring that each sample critically relies on visual content for accurate answers while minimizing data leakage. It comprises 1,500 meticulously selected samples and is organized into six primary sub-benchmarks as follows:
        \begin{itemize}
        \item \textbf{Coarse perception (CP)}, which pertains to the ability to grasp and interpret the overarching features and themes of an image without focusing on minute details,
        \item \textbf{Fine-grained perception (FP)}, which denotes a detailed level of image comprehension that emphasizes the intricate and nuanced aspects of visual content,
        \item \textbf{Instance reasoning (IR)}, which encompasses advanced cognitive abilities aimed at understanding and interpreting individual and collective object attributes and their interrelations within an image,
        \item \textbf{Logical reasoning (LR)}, which involves a sophisticated framework of cognitive processes designed to interpret, deduce, and infer conclusions from visual content through a structured approach to logic and reasoning,
        \item \textbf{Science \& technology (ST)}, which includes a comprehensive framework for the application and integration of knowledge across a wide range of scientific and technological domains,
        \item \textbf{Math (MA)}, which is a fundamental pillar of logical and analytical reasoning and includes a broad spectrum of skills essential for understanding, applying, and interpreting quantitative and spatial information.
        \end{itemize}
    
    \item \textbf{MathVerse}~\citep{zhang2024mathverse} assesses the capabilities of Multi-modal Large Language Models (MLLMs) in visual mathematical reasoning, particularly their ability to understand visual diagrams and mathematical expressions. This dataset is categorized into three primary areas: plane geometry, solid geometry, and functions, and further detailed into twelve types like length and area, encompassing 2,612 visual mathematical challenges.

    To investigate how MLLMs process visual diagrams in mathematical reasoning, the creators of MathVerse developed six distinct versions of each problem, each version presenting different levels of multi-modal information. They initially established three specific classifications for the text content within the problems:
    \begin{itemize}
        \item \textit{Descriptive Information}, which includes content that is directly visible and explicitly depicted in the diagrams,
        \item \textit{Implicit Property}, which encompasses details that demand a more advanced visual perception yet less mathematical knowledge to interpret from the diagram,
        \item \textit{Essential Condition}, which pertains to crucial numerical or algebraic data necessary for solving the problem that cannot be inferred solely from the visual diagram.
    \end{itemize}

    Based on these categories, to thoroughly assess the true visual understanding capabilities of MLLMs and their utility in multi-modal mathematical contexts, the researchers created six versions or sub-benchmarks of each problem in MathVerse, described as follows:
    \begin{itemize}
        \item \textbf{Text dominant (TD)} version, which preserves all textual elements, including the three textual categories and the main question, prompting MLLMs to primarily depend on textual information.
        \item \textbf{Text lite (TL)} version reduces the \textit{Descriptive Information} from the Text dominant version, promoting reliance on the diagram for elementary data.
        \item \textbf{Text only (TO)} version removes the visual elements entirely, focusing on textual content to discern where MLLMs predominantly derive contextual information for problem solving.
        \item \textbf{Vision intensive (VI)} further excludes \textit{Implicit Property} from the Text lite version, urging MLLMs to intensify their visual analysis to gather essential cues for mathematical reasoning.
        \item \textbf{Vision dominant (VD)}, evolving from the Text lite version, omits \textit{Essential Condition} from the textual information and instead visually annotates these details in diagrams, compelling MLLMs to identify and accurately link these essential conditions solely through visual examination.
        \item \textbf{Vision only (VO)} eliminates all textual descriptions, presenting the problem exclusively through visual means and challenging MLLMs to decode and identify mathematical queries purely based on visual data, serving as the ultimate test of their visual reasoning skills in mathematics.
    \end{itemize}

    \item \textbf{VisualWebBench}~\citep{liu2024visualwebbench} assesses the capabilities of multimodal large language models (MLLMs) specifically in the web domain. It is designed to address the lack of a comprehensive benchmark that evaluates the unique characteristics of web pages and measures fine-grained abilities such as OCR, understanding, and grounding in text-rich and interactive web environments. It covers a wide range of domains, including science, travel, sports, engineering, and government, and tasks such as captioning, WebQA, heading OCR, element grounding, and action prediction, containing a total of 1,534 instances.
\end{itemize}

\clearpage
\section{\troltitle TroL Generation Quality}
\label{app:B}
\begin{figure}[h!]
    \centering
    \includegraphics[width=\textwidth]{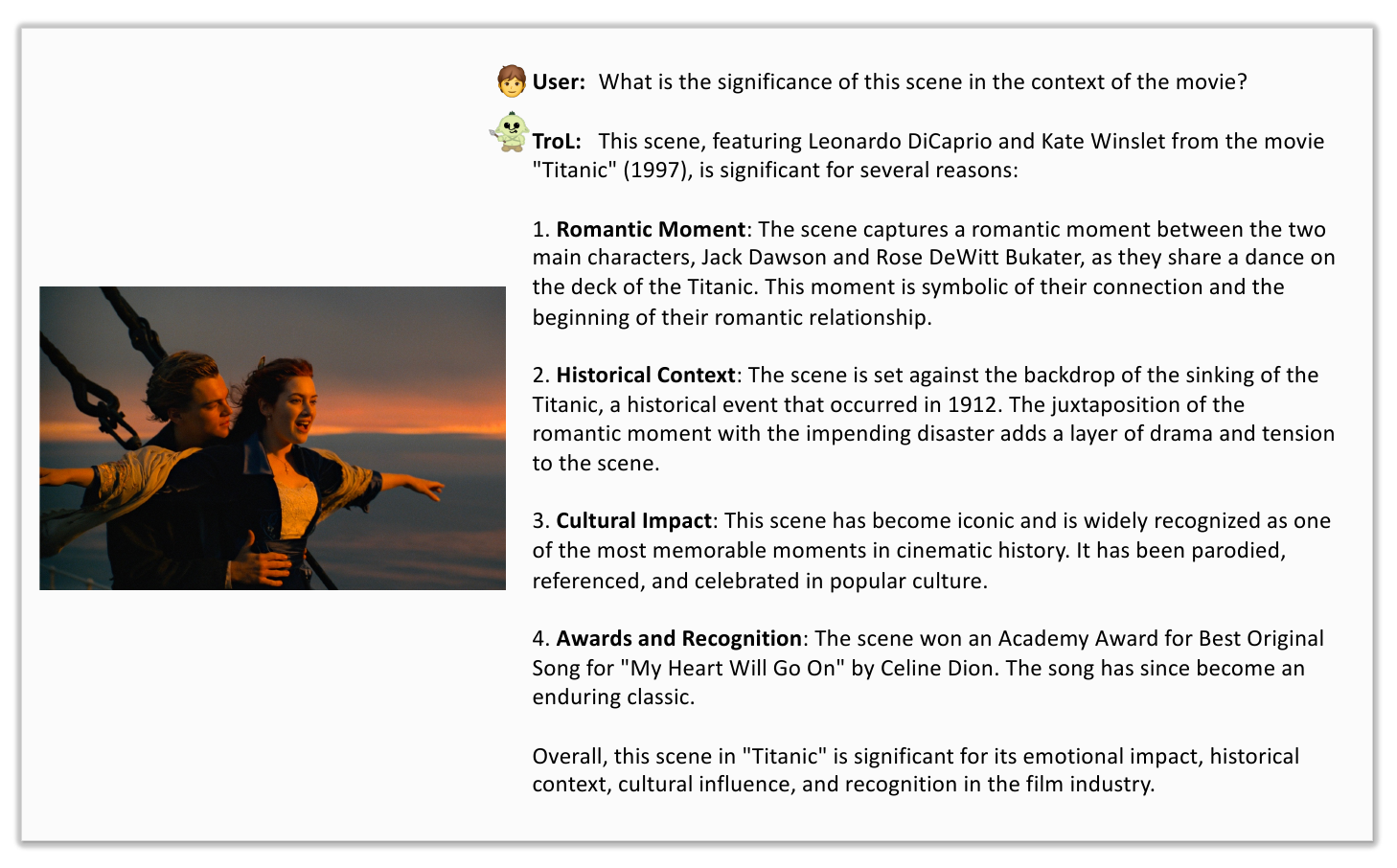}
\end{figure}

\begin{figure}[h!]
    \centering
    \includegraphics[width=\textwidth]{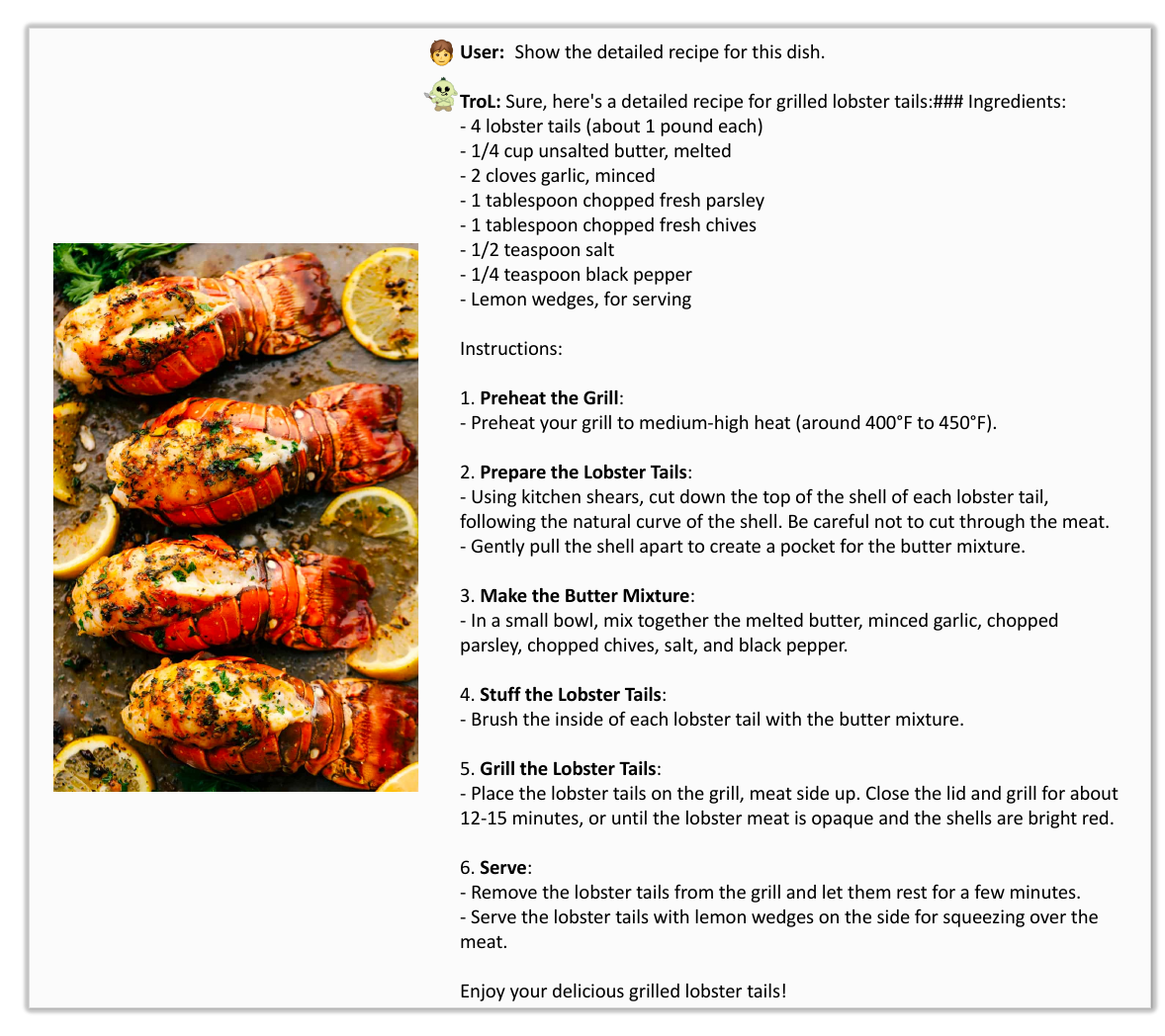}
\end{figure}

\begin{figure}[h!]
    \centering
    \includegraphics[width=\textwidth]{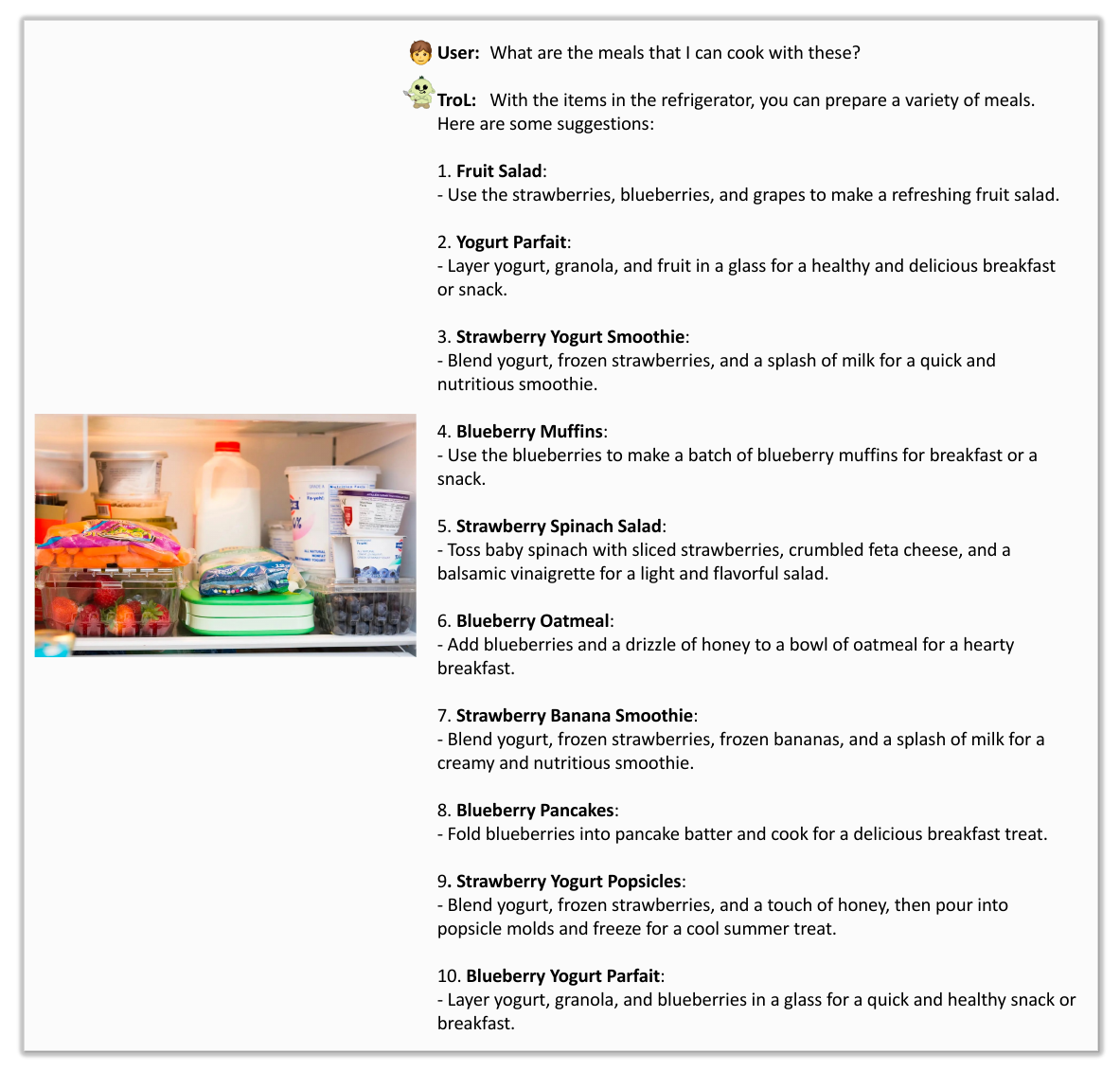}
\end{figure}

\begin{figure}[h!]
    \centering
    \includegraphics[width=\textwidth]{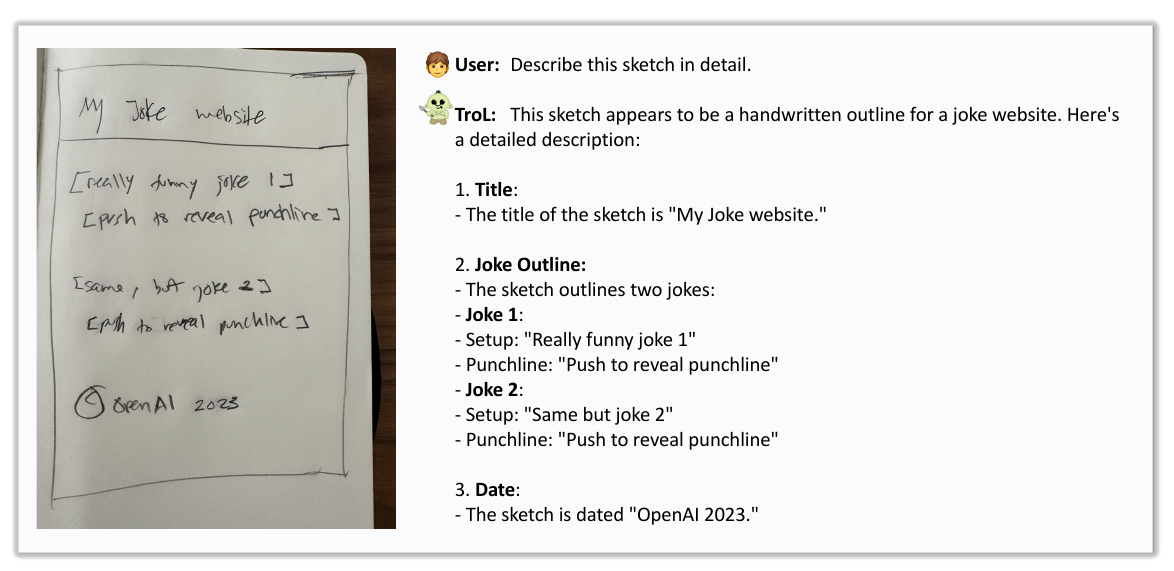}
\end{figure}

\begin{figure}[h!]
    \centering
    \includegraphics[width=\textwidth]{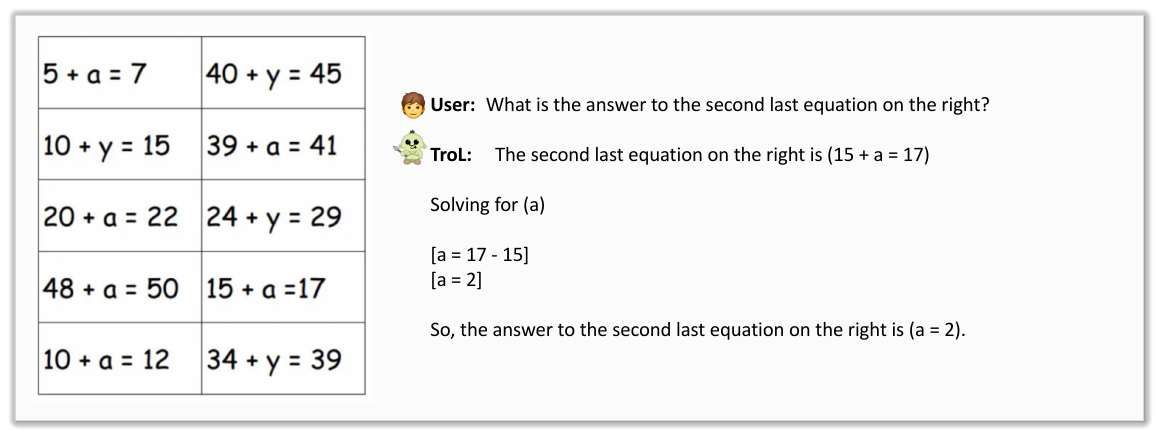}
\end{figure}

\begin{figure}[h!]
    \centering
    \includegraphics[width=\textwidth]{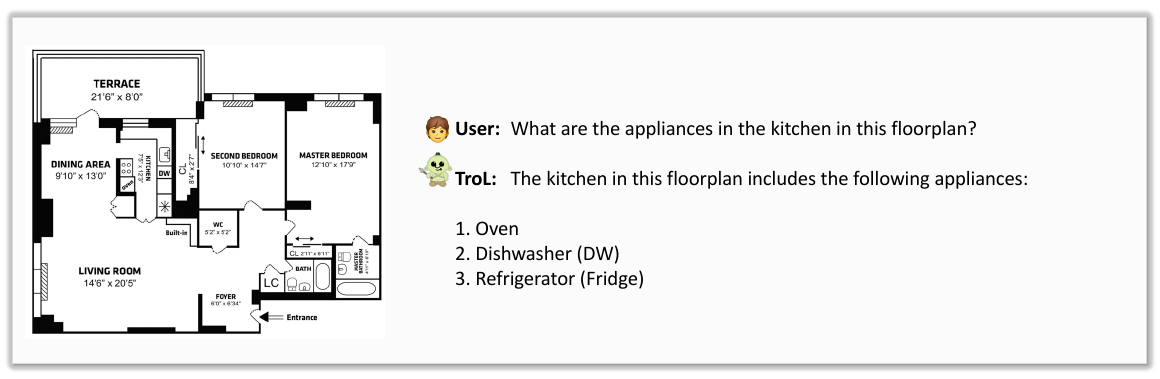}
\end{figure}

\begin{figure}[h!]
    \centering
    \includegraphics[width=\textwidth]{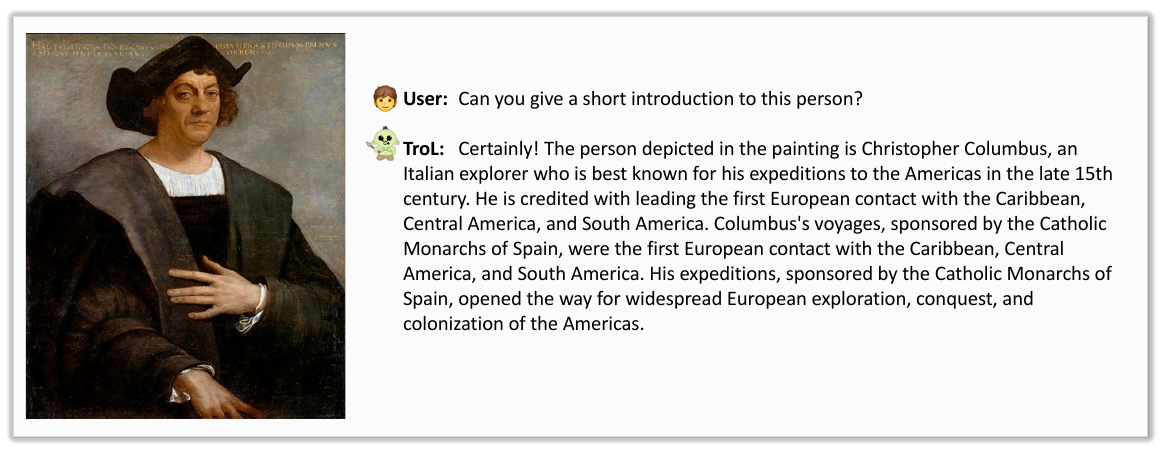}
\end{figure}

\begin{figure}[h!]
    \centering
    \includegraphics[width=\textwidth]{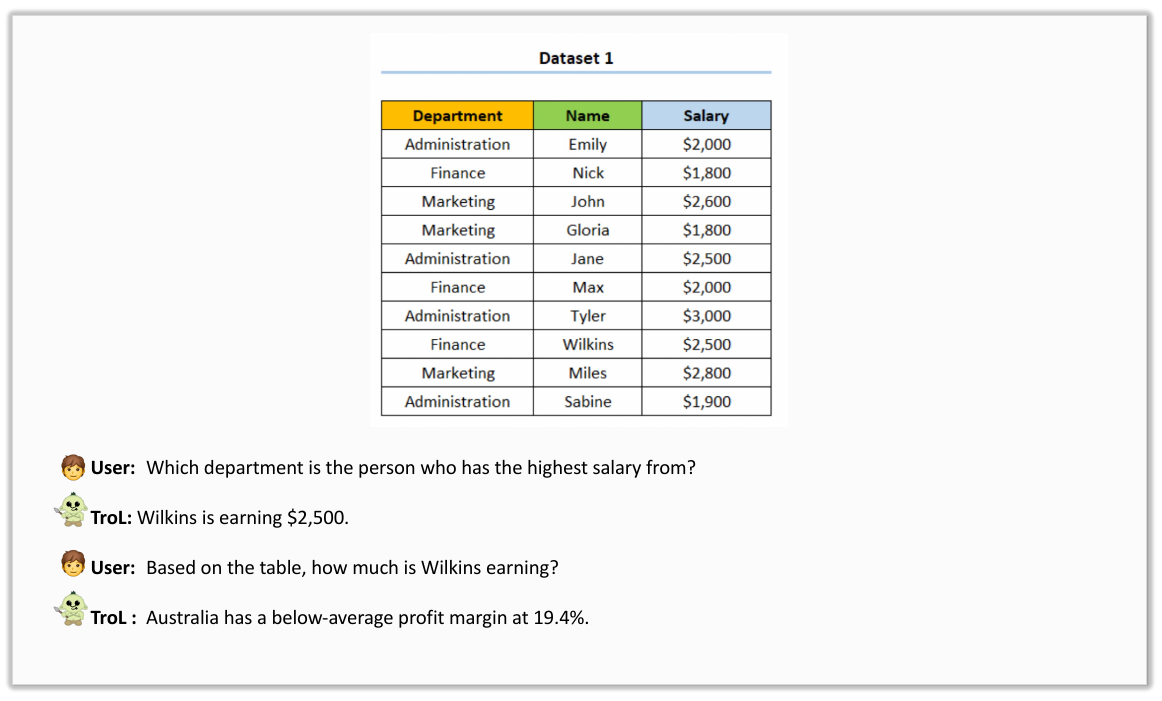}
\end{figure}

\begin{figure}[h!]
    \centering
    \includegraphics[width=\textwidth]{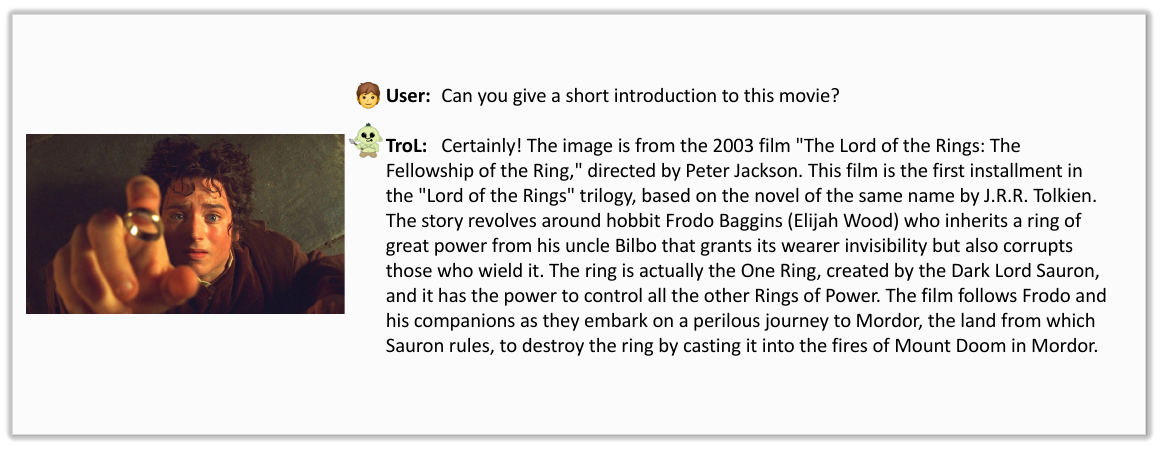}
\end{figure}

\begin{figure}[h!]
    \centering
    \includegraphics[width=\textwidth]{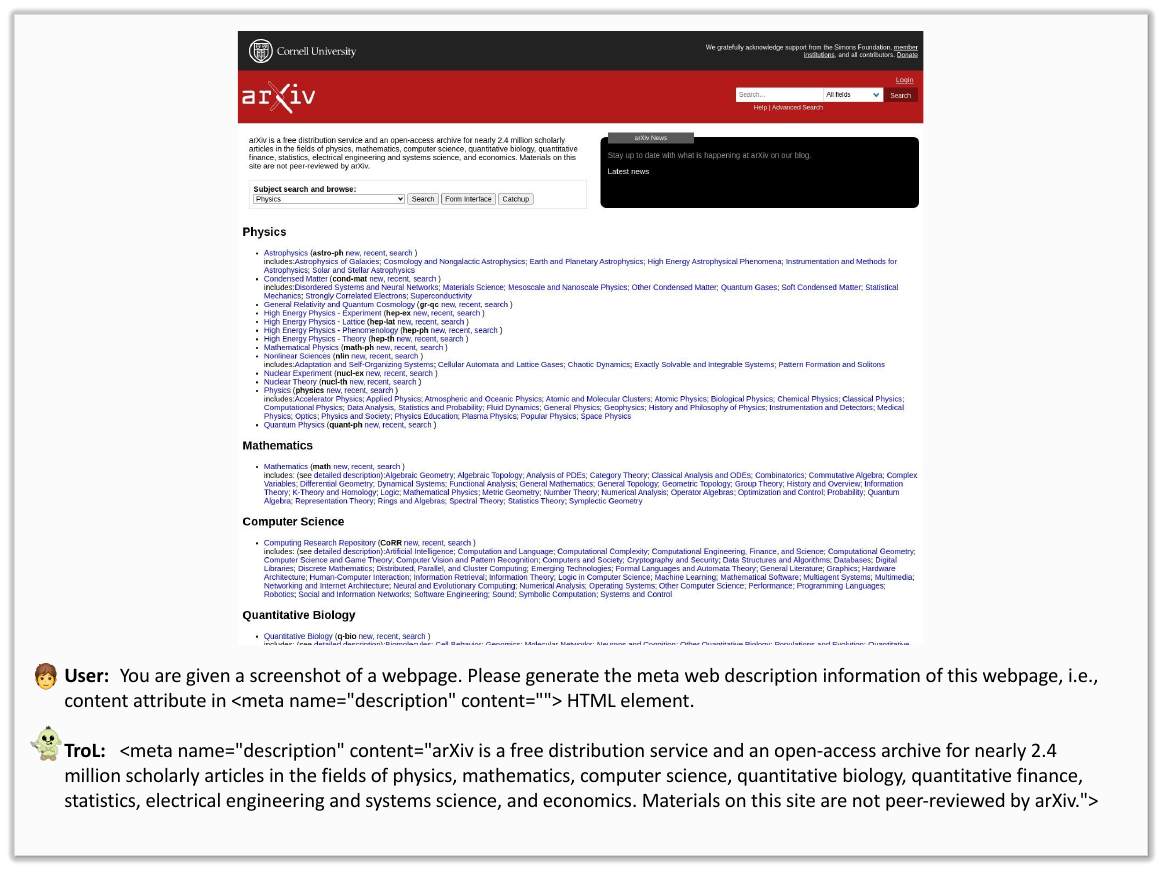}
\end{figure}

\clearpage

\begin{figure}[h!]
    \centering
    \includegraphics[width=\textwidth]{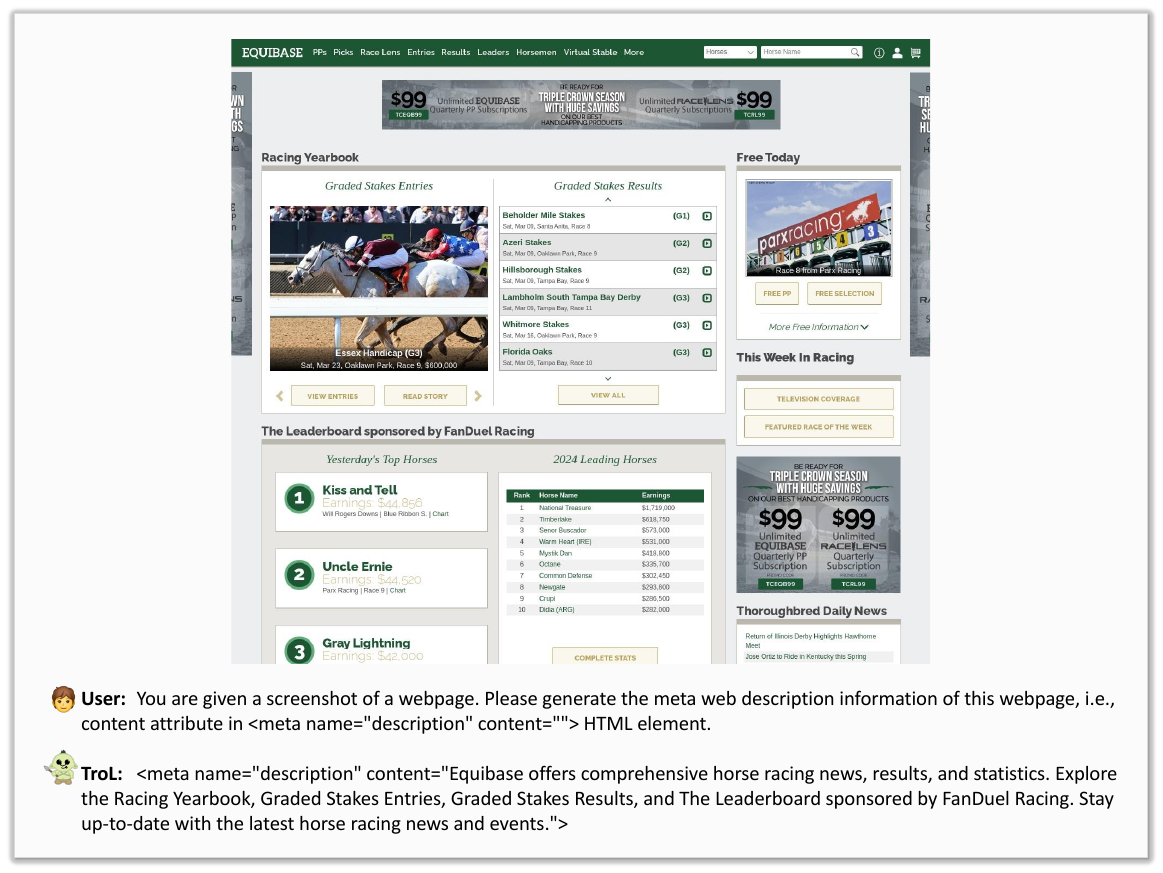}
\end{figure}

\begin{figure}[h!]
    \centering
    \includegraphics[width=\textwidth]{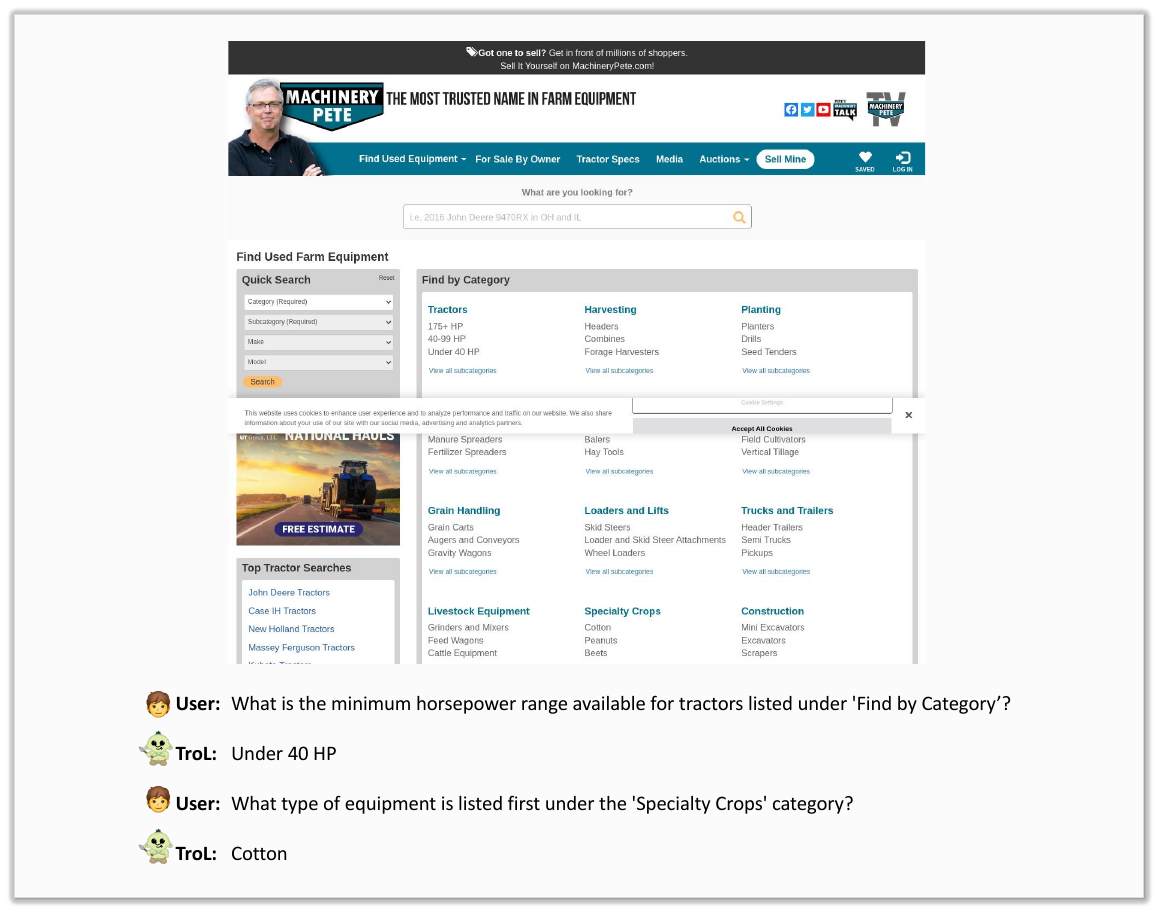}
\end{figure}


\clearpage
\section{Further Ablation Studies}
\label{app:C}

\begin{table}[h!]
\centering
\begin{tabular}{lcccc}
\toprule
Methods                & Added Params & MMStar & MM-Vet & LLaVA$^{\text{W}}$ \\
\midrule
TroL-1.8B w.o. Lay-Trav         & 0                     & 36.0            & 34.6            & 75.0             \\
TroL-1.8B w. MoE (4 Experts, Top k=3) & 402M                  & 39.1            & 38.8            & 80.2       \\
TroL-1.8B w. Lay-Trav           & 49K                   & 45.5            & 45.1            & 87.5             \\
\midrule
TroL-3.8B w.o. Lay-Trav         & 0                     & 37.4            & 35.3            & 81.4             \\
TroL-3.8B w. MoE (4 Experts, Top k=3) & 1.2B                  & 39.9            & 41.1            & 83.4       \\
TroL-3.8B w. Lay-Trav           & 98K                   & 46.5            & 51.0            & 90.8             \\
\midrule
TroL-7B w.o. Lay-Trav           & 0                     & 41.2            & 45.8            & 82.7             \\
TroL-7B w. MoE (4 Experts, Top k=3) & 2.1B                  & 45.6            & 50.5            & 86.6         \\ 
TroL-7B w. Lay-Trav             & 131K                  & 51.3            & 54.7            & 92.8             \\
\bottomrule
\end{tabular}
\caption{Performance comparison of \trol TroL across various benchmarks.}
\label{tab:appC1}
\end{table}

Inspired by the Mixture of Experts (MoE) concept~\citep{shazeer2017, lin2024moe}, we designed the architecture of TroL-Gating and TroL-Mixer. The primary advantage of using MoE is achieving significant performance improvements without adding many physical layers or significantly increasing model sizes.

The paragraph below Equation 2 in \citet{shazeer2017} states: `A simple choice of non-sparse gating function is to multiply the input by a trainable weight matrix and then apply the Softmax function.' Here, the concept of a `gating function' is analogous to TroL-Gating. Similarly, Equation 6 in \citet{lin2024moe} performs a weighted average on the features obtained from each expert module, which is akin to the weighted average operation in TroL-Mixer.

The only difference between traditional MoE and TroL is whether multiple expert modules are physically used. In TroL, the first propagation and the second propagation are implicitly considered as two expert modules in MoE. We believe this iterative propagation introduces a broader range of vision-language knowledge, as layer traversing handles more parameters than MoE. Table~\ref{tab:appC1} below compares the MoE and layer traversing techniques.

Therefore, we conclude that the layer traversing technique embeds more vision-language knowledge than MoE without adding many physical layers or significantly increasing model sizes. This explains why layer traversing performs well across general tasks. Moreover, this approach intuitively matches the human process of retracing answering stream. We hope the layer traversing technique will be regarded as a promising direction for future MoE research.

\begin{table}[h!]
\centering
\begin{tabular}{lcccc}
\toprule
Methods            & Lay-Trav      & MMStar & MM-Vet & LLaVA$^{\text{W}}$ \\
\midrule
DeepSeek-VL-1.3B            & \xmark                 & 39.9            & 34.8            & 51.1             \\
DeepSeek-VL-1.3B            & \cmark                 & 45.9            & 46.2            & 60.5             \\
MiniCPM-2.8B                & \xmark                 & 39.1            & 41.0            & 69.2             \\
MiniCPM-2.8B                & \cmark                 & 47.4            & 50.2            & 80.8             \\
LLaVA-NeXT-7B               & \xmark                 & 40.2            & 43.9            & 72.3             \\
LLaVA-NeXT-7B               & \cmark                 & 49.8            & 52.7            & 84.4             \\
\bottomrule
\end{tabular}
\caption{Performance comparison of various models with and without Lay-Trav across benchmarks.}
\label{tab:appC2}
\end{table}

In Table~\ref{tab:appC2}, we have applied layer traversing to other LLVMs such as DeepSeek-VL-1.3B, MiniCPM-V2-2.8B, and LLaVA-NeXT-7B. This adjustment aims to validate the effectiveness of layer traversing under a fairer comparison setting, using the same TroL visual instruction tuning dataset.

\clearpage 

\begin{table}[h!]
\centering
\begin{tabular}{lccccc}
\toprule
Methods                         & Lay-Trav & Training & MMStar & MM-Vet & LLaVA$^{\text{W}}$ \\
\midrule
TroL-1.8B (Backbone Model)                & \xmark            & \xmark            & 25.1            & 21.4            & 44.6             \\
TroL-1.8B (Backbone Model)                & \cmark            & \cmark            & 24.8            & 15.9            & 36.0             \\
TroL-1.8B                                 & \xmark            & \xmark            & 36.0            & 34.6            & 75.0             \\
TroL-1.8B                                 & \cmark            & \cmark            & 45.5            & 45.1            & 87.5             \\
\midrule
TroL-3.8B (Backbone Model)                & \xmark            & \xmark            & 26.6            & 22.3            & 45.4             \\
TroL-3.8B (Backbone Model)                & \cmark            & \cmark            & 25.2            & 16.0            & 36.1             \\
TroL-3.8B                                 & \xmark            & \xmark            & 37.4            & 43.5            & 78.4             \\
TroL-3.8B                                 & \cmark            & \cmark            & 46.5            & 51.1            & 90.8             \\
\midrule
TroL-7B (Backbone Model)                  & \xmark            & \xmark            & 27.3            & 21.1            & 51.3             \\
TroL-7B (Backbone Model)                  & \cmark            & \cmark            & 24.1            & 15.3            & 37.1             \\
TroL-7B                                   & \xmark            & \xmark            & 41.2            & 45.8            & 82.7             \\
TroL-7B                                   & \cmark            & \cmark            & 51.3            & 54.7            & 92.8             \\
\bottomrule
\end{tabular}
\caption{Performance comparison of \trol TroL with and without Lay-Trav and different training setups.}
\label{tab:appC3}
\end{table}

In Table~\ref{tab:appC3}, we evaluated the experiments you asked in the table below. Without any training, layer traversing technique confuses LLM performances because it never see this kind of features.

\begin{table}[h!]
\centering
\begin{tabular}{lccccc}
\toprule
Methods       & Num of Prop & MMStar & MM-Vet & LLaVA$^{\text{W}}$ \\
\midrule
TroL-1.8B              & 2                    & 45.5            & 45.1            & 87.5             \\
TroL-1.8B              & 3                    & 45.7            & 45.6            & 87.8             \\
TroL-1.8B              & 4                    & 45.8            & 45.9            & 88.0             \\
\midrule
TroL-3.8B              & 2                    & 46.5            & 51.1            & 90.8             \\
TroL-3.8B              & 3                    & 46.6            & 51.4            & 91.1             \\
TroL-3.8B              & 4                    & 46.9            & 51.8            & 91.2             \\
\midrule
TroL-7B                & 2                    & 51.3            & 54.7            & 92.8             \\
TroL-7B                & 3                    & 51.7            & 55.0            & 93.2             \\
TroL-7B                & 4                    & 52.0            & 55.2            & 93.3             \\
\bottomrule
\end{tabular}
\caption{Performance comparison of \trol TroL across different numbers of propositions.}
\label{tab:appC4}
\end{table}

The reason we use the second propagation instead of the third propagation is that marginal improvements are observed when using more than two propagation. Table~\ref{tab:appC4} shows the performance across the propagation numbers from 2 to 4.

\begin{table}[h!]
\centering
\begin{tabular}{lcccc}
\toprule
Methods      & Avg Time Ratio & MMStar & MM-Vet & LLaVA$^{\text{W}}$ \\
\midrule
TroL-1.8B (Question)            & 1.0                    & 45.5            & 45.1            & 87.5             \\
TroL-1.8B (Question-Answer)     & 4.9                    & 46.3            & 47.2            & 89.2             \\
\midrule
TroL-3.8B (Question)            & 1.0                    & 46.5            & 51.1            & 90.8             \\
TroL-3.8B (Question-Answer)     & 5.1                    & 47.9            & 53.4            & 92.1             \\
\midrule
TroL-7B (Question)              & 1.0                    & 51.3            & 54.7            & 92.8             \\
TroL-7B (Question-Answer)       & 5.7                    & 52.5            & 56.2            & 94.3             \\
\bottomrule
\end{tabular}
\caption{Performance comparison of \trol TroL across question and question-answer setups.}
\label{tab:appC5}
\end{table}

The primary reason for not applying layer traversing to the answer part is the huge increase in answering time complexity. Fortunately, turning layer traversing on or off for the answer part during inference shows a little significant performance gap in Table~\ref{tab:appC5}, therefore it is better for us to deal with only question part in efficient inference.

\begin{table}[h!]
\centering
\begin{tabular}{lccccc}
\toprule
Methods                          & Lay-Trav in Training & MMStar & MM-Vet & LLaVA$^{\text{W}}$ \\
\midrule
TroL-1.8B                                 & \xmark                        & 36.0            & 34.6            & 75.0             \\
TroL-1.8B (Nothing)                       & \cmark                        & 42.1            & 41.0            & 82.8             \\
TroL-1.8B (Question)                      & \cmark                        & 45.5            & 45.1            & 87.5             \\
TroL-1.8B (Question-Answer)               & \cmark                        & 46.3            & 47.2            & 89.2             \\
\midrule
TroL-3.8B                                 & \xmark                        & 37.4            & 43.5            & 78.4             \\
TroL-3.8B (Nothing)                       & \cmark                        & 43.8            & 46.3            & 86.3             \\
TroL-3.8B (Question)                      & \cmark                        & 46.5            & 51.1            & 90.8             \\
TroL-3.8B (Question-Answer)               & \cmark                        & 47.9            & 53.4            & 92.1             \\
\midrule
TroL-7B                                   & \xmark                        & 41.2            & 45.8            & 82.7             \\
TroL-7B (Nothing)                         & \cmark                        & 47.0            & 50.8            & 88.9             \\
TroL-7B (Question)                        & \cmark                        & 51.3            & 54.7            & 92.8             \\
TroL-7B (Question-Answer)                 & \cmark                        & 52.5            & 56.2            & 94.3             \\
\bottomrule
\end{tabular}
\caption{Performance comparison of \trol TroL with Lay-Trav in training and various configurations.}
\label{tab:appC6}
\end{table}

In Table~\ref{tab:appC6}, thanks to layer traversing together with visual instruction tuning, we observed that vision language performance improves even when question and answer traversing are turned off.

\end{document}